\documentclass{article} 
\PassOptionsToPackage{table,xcdraw}{xcolor} 
\RequirePackage{xcolor}
\usepackage{iclr2025_conference,times}

\usepackage{amsmath,amsfonts,bm}

\def\eqref#1{equation~\ref{#1}}

\def\1{\bm{1}}

\DeclareMathAlphabet{\mathsfit}{\encodingdefault}{\sfdefault}{m}{sl}
\SetMathAlphabet{\mathsfit}{bold}{\encodingdefault}{\sfdefault}{bx}{n}

\usepackage{hyperref}
\usepackage{url}
\usepackage{algorithm}
\usepackage{algpseudocode}
\usepackage{amsthm}
\usepackage{booktabs}
\usepackage{multirow}
\usepackage{graphicx}
\usepackage{adjustbox}
\usepackage{wrapfig}
\usepackage{subfigure}  
\usepackage{xspace}
\usepackage{titletoc}
\usepackage{comment}
\usepackage{ulem}
\usepackage[font=small]{caption}

\newcommand{\atk}{DocMIA\xspace}

\newcommand{\atkFL}{FL\xspace}
\newcommand{\atkFLLoRA}{FLLoRA\xspace}
\newcommand{\atkIG}{IG\xspace}

\title{DocMIA: Document-Level Membership Inference Attacks against DocVQA Models}

\author{Khanh Nguyen$^{1}$~~~~~Raouf Kerkouche$^{2}$\thanks{The corresponding author.}~~~~~Mario Fritz$^{2}$~~~~~Dimosthenis Karatzas$^{1}$ \\
$^1${Computer Vision Center, Universitat Aut\`onoma de Barcelona}\\ $^2${CISPA Helmholtz Center for Information Security}\\
\texttt{\{knguyen,dimos\}@cvc.uab.es}\\ \texttt{\{raouf.kerkouche,fritz\}@cispa.de}
}

\theoremstyle{definition}
\newtheorem{definition}{Definition}[section]

\iclrfinalcopy 
\begin{document}

\maketitle

\begin{abstract}
Document Visual Question Answering (DocVQA) has introduced a new paradigm for end-to-end document understanding, and quickly became one of the standard benchmarks for multimodal LLMs. Automating document processing workflows, driven by DocVQA models, presents significant potential for many business sectors. However, documents tend to contain highly sensitive information, raising concerns about privacy risks associated with training such DocVQA models. One significant privacy vulnerability, exploited by the membership inference attack, is the possibility for an adversary to determine if a particular record was part of the model's training data. In this paper, we introduce two novel membership inference attacks tailored specifically to DocVQA models. These attacks are designed for two different adversarial scenarios: a white-box setting, where the attacker has full access to the model architecture and parameters, and a black-box setting, where only the model's outputs are available. Notably, our attacks assume the adversary lacks access to auxiliary datasets, which is more realistic in practice but also more challenging. Our unsupervised methods outperform existing state-of-the-art membership inference attacks across a variety of DocVQA models and datasets, demonstrating their effectiveness and highlighting the privacy risks in this domain.
\end{abstract}

\section{Introduction}

Automated document processing fuels a significant number of operations daily, ranging from fintech and insurance procedures to interactions with public administration and personal record keeping. Up until a few years ago, document processing services relied on template-based information extraction models, which were created ad-hoc for each client. Although these approaches allowed for good control of client data and could be extended to new documents with a few examples, they were limited in scalability and difficult to maintain. Consequently, the introduction of Document Visual Question Answering (DocVQA) \citep{mathew2020document} in 2019 has resulted in a paradigm shift in document processing services, enabling end-to-end generic solutions to be applied in this domain. DocVQA leverages multi-modal large language models to streamline business workflows and provide clients with novel ways to interact with the document processing pipeline. 


However, as cloud-based DocVQA solutions become more prevalent, significant privacy risks emerge, particularly concerning the potential leakage of sensitive information through model vulnerabilities.
Indeed, during the training of a DocVQA model, \textit{each document can have several associated question-answer pairs}, with each pair considered a unique data point. As a result, a single document can appear multiple times, which significantly raises the risks associated with privacy vulnerabilities. This repeated exposure enhances the likelihood of the model memorizing specific details, thereby increasing the potential for data leakage during privacy attacks.
Furthermore, scanned document images often have high resolutions necessary for posterior analysis, but need to be rescaled for processing by image encoders, potentially rendering content unreadable. To mitigate this issue, many DocVQA models~\citep{huang2022layoutlmv3,tang2023unifying} utilize a dual representation of the document, comprising both a reduced-scale image and OCR-recognized text. This approach introduces further challenges, as sensitive information may leak through multiple modalities.

\begin{figure}[t]
    \centering
    \includegraphics[width=\textwidth]{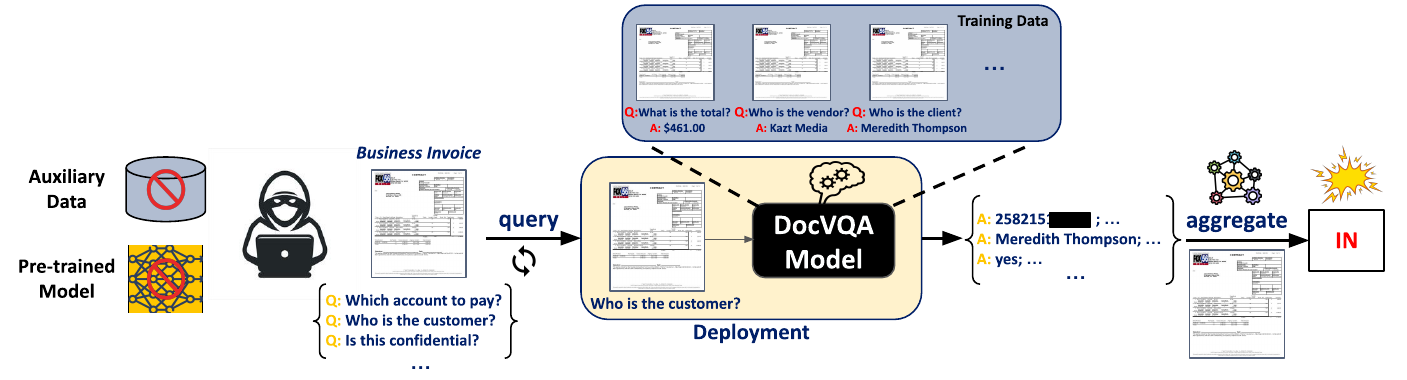}
    \caption{\textbf{The General Scheme of Document-level Membership Inference Attacks.} \textit{Training:} A DocVQA model is trained on a dataset of documents, each associated with \textbf{multiple questions/answers}. \textit{Deployment:} An adversary exploits this structure by querying the model with questions related to a target document. By aggregating the model's responses, the adversary can infer the membership of the document in the training set.}
    \label{fig:teaser}
\vspace{-0.2in}
\end{figure}

Membership inference attacks (MIAs) are among the most prominent techniques for assessing privacy vulnerabilities in machine learning models. These attacks enable an adversary to determine whether a specific data point is included in the training dataset. However, there is limited research on membership inference risks in the context of multi-modal models. Among the few studies, \citet{ko2023practical,hu2022m} utilize powerful pre-trained models on large datasets to construct an aligned embedding space for the two modalities—image as input and text as output—allowing for the inference of membership information. Unfortunately, the reliance on these pre-trained models poses challenges for document-based tasks, particularly in DocVQA scenarios, where an alignment model capable of aligning the (document, question) as input and the answer as output is currently unavailable. Recently, \citet{tito2024privacy} introduced a provider-level MIAs against DocVQA models aimed at determining whether a \textit{provider (group)} that may supply multiple invoice documents is part of the training set. In contrast, our research focuses on membership information at a finer granularity, specifically targeting the inference of whether a \textit{single document} is included in the training dataset. Current MIA solutions that exploit standard features such as output logits, probabilities, or loss are difficult to adapt to the DocVQA context, where outputs are generated in an auto-regressive manner. Additionally, legal constraints surrounding copyright and private information complicate centralized model training, making it challenging to create auxiliary datasets that capture the variability and richness of real-world data. As a result, shadow training of proxy models becomes infeasible.

In this work, we take a structured approach to privacy testing for DocVQA models. We design a novel Document-level Membership Inference Attack (\atk) that deals with the multiple occurrences of the same document in the training set, as demonstrated in Figure \ref{fig:teaser}. To address the challenge of extracting typical metrics (e.g. logit-based) from auto-regressive outputs, we propose a new method based on model optimisation for individual samples that generates discriminative features for \atk. We design attacks both for white-box and black-box settings without requiring auxiliary datasets. In the black-box setting, we propose an alternative knowledge transfer mechanism from the attacked model to a proxy. Evaluating our attacks on three multi-modal DocVQA models and two datasets, we achieve state-of-the-art performance against multiple baselines.

To summarize, we make the following contributions:
\begin{enumerate}
    \item We present \atk, the first Document-level Membership Inference Attacks specifically targeting multi-modal models for DocVQA.
    \item We introduce two novel auxiliary data-free attacks for both white-box and black-box settings, leveraging novel discriminative metrics for \atk.
    \item We explore three distinct approaches to quantify these metrics: vanilla layer fine-tuning  (\atkFL), fine-tuning layer with LoRA \citep{hu2021lora} (\atkFLLoRA), and image gradients (\atkIG).
    \item Our attacks\footnote{Code is available at ~\url{https://github.com/khanhnguyen21006/mia_docvqa}}, evaluated on two DocVQA datasets across three different models, outperform existing state-of-the-art membership inference attacks as well as baseline attacks.
\end{enumerate}

\section{Related Work}
\label{sec:related_work}
\vspace{-0.15in}

\paragraph{Membership Inference Attack.} 
Membership inference attacks have been extensively explored in various applications to highlight privacy vulnerabilities in deep neural networks or to audit model privacy~\citep{shokri2017membership}. These attacks are categorized into two types: white-box and black-box settings. In white-box settings, the adversary has full access to the target model's internal parameters and computations~\citep{carlini2022membership, yeom2018privacy, nasr2019comprehensive, rezaei2021difficulty, sablayrolles19a, li2021membership}, enabling the use of informative features like loss values, logits, and gradient norms. Conversely, in black-box settings, the adversary is limited to the model's outputs, such as predicted labels or confidence scores~\citep{choquette2021label, shokri2017membership, salem2018ml, sablayrolles19a, song2021systematic, hui2021practical}. The literature indicates that white-box attacks tend to be more effective due to the availability of richer features~\citep{song2019privacy,nasr2019comprehensive}. In this paper, we propose tailored attacks for both settings, considering a more challenging scenario where the adversary lacks an auxiliary dataset --which is used to train shadow models that mimic the behavior of the target model and are subsequently exploited to enhance attack performance-- and is restricted to a limited number of queries. Regarding gradient-based membership inference attacks, research on using gradients as features has been limited. \citet{nasr2019comprehensive} leveraged the $L2$-norm of gradients with respect to model weights for membership inference. \citet{rezaei2021difficulty} suggested using the distance to the decision boundary as a metric but found it ineffective for this purpose. In contrast, we introduce novel strategies called \atkFL, \atkFLLoRA, and \atkIG, demonstrating that the $L2$-norm of the cumulative gradient—computed using these methods—provides a robust signal for membership inference. While \citet{maini2021dataset} and \citet{li2021membership} also explored distance metrics, but from input points for membership inference in image classification tasks, their approaches lack scalability and applicability in our context, which involves larger-scale models with a wider vocabulary of tokens.

\paragraph{Membership Inference Attack Against Multi-modal Models.} Research works into the privacy vulnerabilities of multi-modal models is still in its early stages. Recently, \citet{tito2024privacy,pinto24a} proposed reconstruction attacks that exploit DocVQA model memorization to recover hidden values in documents. They black out specific target values in documents and query the model with questions about the modified documents. Since the model memorizes training data, it often reconstructs the hidden target values. \citet{tito2024privacy} also introduced a membership attack against DocVQA models to infer whether a document provider, with multiple documents, is included in the training dataset. However, as far as we know, no research has yet explored membership inference attacks at document-level granularity. Additionally, \citet{ko2023practical,hu2022m} leverage powerful \textit{pre-trained models} on large datasets to create an aligned embedding space for the two modalities to infer membership. Unfortunately, the reliance on these pre-trained models introduces difficulties for document-based tasks, especially DocVQA, where an appropriate alignment model for aligning (document, question) inputs to corresponding answers is not yet available. Furthermore, the success of both attacks hinges on the availability of \textit{auxiliary datasets} leveraged by the adversary, which are key to executing the attack effectively. In this paper, we present two membership inference attacks specifically tailored to tackle the unique characteristics of DocVQA models.

\section{Background}

\subsection{Document-based Visual Question Answering}
DocVQA is a multi-modal task where natural language questions are posed based on the content of document images. Notably, it establishes a unified query-response framework applicable across various document understanding tasks, such as document classification and information extraction.

Formally, the DocVQA task is defined as follows: given a question-answer pair $(q, a)$ related to a document image $x$, the method $\mathcal{F}$ must generate an answer $\hat{a}=\mathcal{F}(x,q)$ such that $\hat{a}$ closely matches the correct answer $a$. More concretely, given $D_t = \{(x_i,q_i,a_i)\}^{N_t}_{i=1}$ as a set of valid training examples, a model $\mathcal{F}$, parameterized by $\theta$, is trained to maximize the conditional log-likelihood:
\begin{equation}
    \mathcal{L}(\theta) =-\log{p_{\theta} (a_i|x_i,q_i)}
\label{eq:training_loss}
\end{equation}

Standard metrics for DocVQA include Accuracy (ACC) and Normalized Levenshtein Similarity (NLS) \citep{biten2019scene}, which measure the similarity between the predicted and correct answer:
\begin{equation}
    \textsc{ACC} = \displaystyle \1_\mathrm{\hat{a} = a};
    \quad
    \textsc{NLS} = 
    \begin{cases}
        1 - \text{NL}(\hat{a}, a) & \text{if } \textsc{NL}(\hat{a}, a) < 0.5, \\
        0 & \text{if } \textsc{NL} \geq 0.5
    \end{cases}
\end{equation}
where $\textsc{NL}(\cdot,\cdot)$ denotes the normalized Levenshtein distance.

In the following sections, for clarity, we often omit the data example index $i$ from the notation.

\subsection{Document-level Membership Inference Attack}
Membership inference attacks~\citep{shokri2017membership} exploit privacy vulnerabilities to determine if a specific data point was included in the training set of a machine learning model. We extend this definition to the Document-level MIA, which is particularly suited in the DocVQA context. 

Given access to a trained DocVQA model $\mathcal{F}$ and a document $x$ drawn from its data distribution $\mathcal{D}$, along with a set of question-answer pairs $Q=\{(q_i,a_i)\}_{i=1}^{M}$ related to the information in the document, an adversary $\mathcal{A}$ designs a decision rule $f_{\mathcal{A}}(x,Q;\mathcal{F})$  to classify the membership status of $x$, aiming for $f_{\mathcal{A}}(x,Q;\mathcal{F})=1$ if $x$ is a member of the training set, otherwise a non-member. It is important to note that the adversary is focused solely on \textit{the membership of the document $x$}, rather than the entire DocVQA \textit{data point} $(x,q,a)$, which is typically the target of prior MI attacks. Moreover, since a single document is associated with many question-answer pairs, this allows the adversary to query the same document using multiple questions for various pieces of information.

\section{\atk against DocVQA models}
\vspace{-0.1in}

In Section~\ref{sec:threat_model}, we elaborate on the threat model relevant to \atk on two scenarios: \textit{white-box} and \textit{black-box} access. We first explain our intuition behind our optimization-based attacks in the white-box setting (Section~\ref{sec:docmia_whitebox}), then adapt this approach to our black-box attacks (Section~\ref{sec:docmia_blackbox}).
\vspace{-0.1in}
\subsection{Threat Model}
\label{sec:threat_model}
\atk can be either a useful or harmful tool in various real-world scenarios. On the positive side, \atk can act as a privacy auditing tool. For instance, in legal document processing, law firms may use these attacks to evaluate whether proprietary or confidential documents, such as contracts or court filings, were included in model training, thereby identifying potential privacy risks.
Conversely, \atk can be maliciously leveraged. As an example, a business competitor could exploit these attacks on an invoice-processing system to infer the presence of specific invoices in the training data, exposing confidential business relationships and leading to risks such as supplier poaching.

In both scenarios, we assume that the adversary aims to infer membership information for a set of documents, determining whether each document is included in the training dataset. These documents may or may not be part of the target model's training data.
Crucially, we further assume \textit{the adversary lacks access to an auxiliary data} $D_{\text{aux}}$ that reflects the characteristics of these documents. This assumption is realistic, as obtaining real-world documents at scale is often prohibitively difficult due to their confidential nature and regulatory restrictions. Consequently, this negates the application of MI attack techniques that require training shadow models~\citep{shokri2017membership,carlini2022membership}. Even if auxiliary documents were available, training numerous shadow document-based models—typically designed with a large number of parameters—would be prohibitively expensive.

Based on the previous examples, we refer to the owner of the document model as \textit{the trainer} and the law firms or competitors as \textit{the adversary}. Given the document distribution $\mathcal{D}$, the trainer trains a document-based model $\mathcal{F}_t$ with private access to $D_t \sim \mathcal{D}$, following a training algorithm $\mathcal{T}$, that defines the model architecture, optimization process, and related details. The adversary owns the set of sensitive documents $D_{\text{test}} \sim \mathcal{D}$, where $D_t \cap D_{\text{test}} \ne \emptyset$, $\vert D_{\text{test}} \vert =N_\text{test}$; but does not know which documents are in $D_t$. Given a document $x \in D_{\text{test}}$ with a set of related queries $Q=\{(q_i,a_i)\}_{i=1}^{M}$, the adversary's goal is to determine whether $x \in D_t$ or $x \notin D_t$. We formulate two attack settings, which specify the adversarial knowledge about the model $\mathcal{F}_t$ and its data distribution $\mathcal{D}$:

\textbf{White-box Setting}. In this scenario, the adversary has full access to the internal workings of the target model, including the model's architecture, weights, gradients from any further training and other internal details. However, the adversary does not have access to the training algorithm $\mathcal{T}$.

\textbf{Black-box Setting}. Here, the adversary can only interact with the target model through an API, which only returns a prediction $\hat{a}$ for each question $q$ on $x$. In addition, the adversary is constrained by a limited number of queries. As in the white-box setting, the adversary has no information about $\mathcal{T}$. This setting reflects the most challenging case~\citep{nasr2019comprehensive,song2019privacy}.

We assume the adversary has full knowledge of the DocVQA task to train the model, including the training objective, document type and exact training questions. 
This assumption is reasonable, as task-level information such as document type, is often publicly available to guide users, making it accessible to adversaries.
The assumption of the exact question knowledge is also plausible, as \textit{an adversary can approximate questions based on the document type}. Further discussion of this assumption and experiments in the setting without exact questions are provided in Appendix~\ref{sec:impact_question_knowledge}.
\vspace{-0.1in}

\subsection{White-box DocMIA}
\label{sec:docmia_whitebox}
In the white-box setting, where the adversary has access to the trained model, shadow training is impractical due to the lack of auxiliary data and high computational cost. To this end, our strategy is to develop unsupervised \textit{metric-based} attacks \citep{hu2022membership}. For each document, we extract a set of features from individual question-answer pairs and aggregate them across all pairs. We then cluster the resulting feature vectors to distinguish member from non-member documents. 
A key challenge is to design discriminative features, as standard metrics (e.g., logit and loss) may be ineffective in this setting (Section \ref{sec:evaluation}). To address this, we propose new features that enhance the informativeness of our membership inference vectors.

\subsubsection{Optimization-Based Discriminative Features}
In this section, we introduce two novel discriminative membership features derived from an optimization process for our attacks against DocVQA models.
\begin{wrapfigure}{R}{0.3\textwidth}
\centering
 \adjustbox{width=.3\textwidth,frame=0.01cm 0cm}{\includegraphics{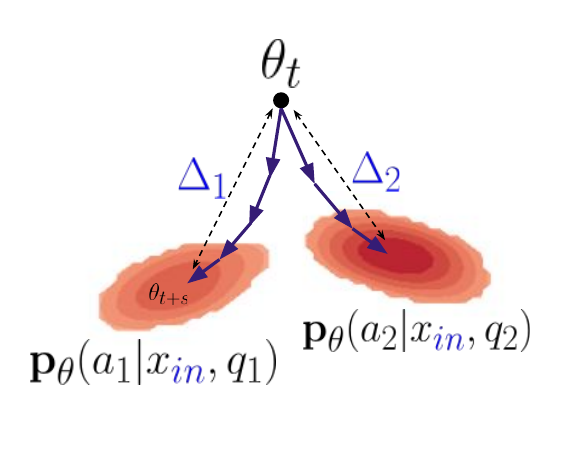}}\hfill
 $x$ is \textbf{\textcolor{blue}{in}} of the training set 
\medskip

\adjustbox{width=.3\textwidth,frame=0.01cm 0cm}{\includegraphics{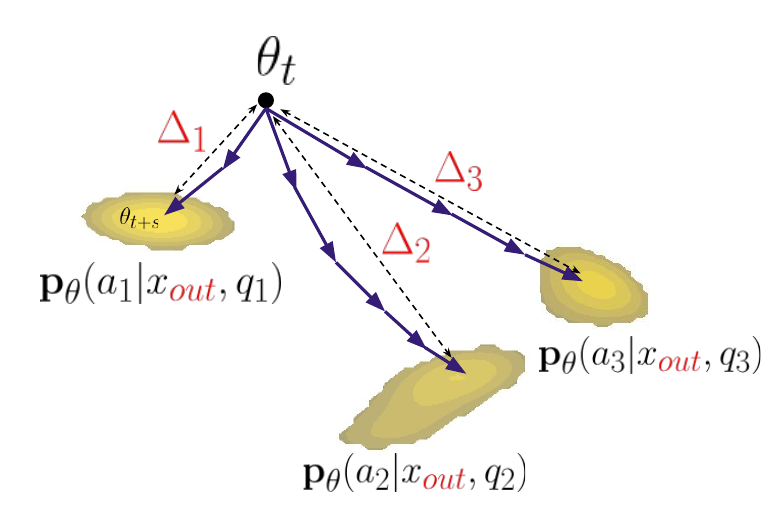}}\hfill
$x$ is \textbf{\textcolor{red}{out}} of the training set 
\caption{\textbf{Visualization of our fine-tuning strategy} in the parameters space. Each contour plot represents the optimization landscape w.r.t each pair $(a_i,q_i)$ from document $x$.  In general, the average $\Delta$ computed on a member document $x_{\textcolor{blue}{\textbf{in}}}$ is smaller than non-member document $x_{\textcolor{red}{\textbf{out}}}$.}

\vspace{-0.3in}
\label{fig:intuition}
\end{wrapfigure}

\textbf{Intuition.} Since DocVQA models are typically trained on multiple question-answer pairs per document, the model parameters likely converge to \textit{minimize the average distance to the ground-truth answers} after training. As a result, fine-tuning the model on one question-answer pair through an iterative process is necessary to extract more reliable membership signals. More importantly, this optimization on training documents may converge faster than to non-training documents, due to the lower generalization error. Figure \ref{fig:intuition} illustrates our reasoning.

We provide a formal definition of the \textit{distance} feature.  
\begin{definition}[\textbf{Optimization-based Distance Feature}] Given a model $\mathcal{F}$ parameterized by $\theta$, let the model be initialized with $\theta_0$. After undergoing a gradient-based optimization process $\mathcal{O}$, the parameters converge to $\theta^{*}$ according to a specified training objective $\mathcal{L}$. The \textit{distance} feature is then defined as the $L2$-norm of the change in parameters:
\begin{equation}
    \Delta(\theta_0, \theta^{*}) = || \theta_0 - \theta^{*} ||_2
\label{eq:distance}
\end{equation}
\end{definition}
This feature measures the difference between the initial parameters $\theta_0$ and the converged parameters $\theta^*$, as an approximation of the optimization trajectory toward the optimal solution. 

Specifically, we fine-tune the target DocVQA model on an individual document/question-answer pair and compute the \textit{distance} required to reach the \textit{optimal} answer. A small average distance indicates the document is likely part of the training set, while a larger distance suggests a non-training document. In addition, the number of optimization steps serves as an orthogonal feature that reflects the efficiency of the optimization process. With an optimal learning rate and a good initialization provided from the target model, optimization for training documents typically converges in fewer steps compared to non-training documents. Consequently, we include both the \textit{distance} and the number of optimization steps in our feature set for white-box attacks.
\begin{algorithm}[tb]
    \footnotesize 
   \caption{\atk Assignment}
   \label{alg:assignment}
    \begin{algorithmic}[1]
       \State {\bfseries Input:} model $\mathcal{F}_{\theta_{t}}$, document $x\in D_{\text{test}}$, question-answer pairs $\{(q_i,a_i)\}_{i=1}^{M}$, utility $\mathcal{U}$, aggregation $\Phi$.
       \State {\bfseries Hyperparameters:} optimizer $\textsc{OPT}$, optimization steps $S$, learning rate $\alpha$, threshold $\tau$.
    
       \For{$i=1$ {\bfseries to} $M$}
       \State \textcolor{blue}{Set ${\theta}.\text{requires\_grad}=\text{True}$ \quad  // Change $\theta$ to: $\theta_L$ or $\textsc{LoRA}(\theta_L)$ or $x$ and Freeze $\theta$.}
       \State Initialize: $s_i = 0, u_{i} = \{\}; \mathit{l}_i \gets 0, \theta_0 \gets {\theta}_{t}$
    
       \While{$s_i < S$}
       \State $u_{i} \gets u_{i} \cup \mathcal{U}({\mathcal{F}}_{\theta}(x, q_i), a_i)$
       \If{$(\mathcal{L}(\theta)-\mathit{l}) < \tau$} 
       break; \quad \textcolor{gray}{// Early stopping}
       \EndIf 
       \State ${\theta} \gets \textsc{OPT}(\alpha, \nabla_{{\theta}}(\mathcal{L}(\theta))$
       \State $\mathit{l}_i \gets \mathcal{L}(\theta), s_i \gets s_i+1$
       \EndWhile
       \State ${\Delta}_{i} \gets \lVert \theta_0-{\theta} \rVert_2$ \quad \textcolor{gray}{// Compute distance metric}
        \EndFor
       \State $\Delta_{M} = \Phi(\Delta_{i={1,\dots,M}}); s_{M}=\Phi({s}_{i={1,\dots,M}}); u_{M}=\Phi({u}_{{i={1,\dots,M}}})$ \quad \textcolor{gray}{// Aggregating over $M$ questions}
       \State {\bfseries Output:} ${F}_{x} = [\Delta_{M},s_{M},u_{M}]$  \quad \textcolor{gray}{// Assign membership feature vector}
    \end{algorithmic}
\end{algorithm}

\vspace{-0.05in}
\subsubsection{Methodology}
We now formally present our attack strategy, assuming white-box access to the target model $\mathcal{F}_t$.
For any document $x \in D_{\text{test}}$ and a set of question-answer pairs $Q$, the goal is to assign a features descriptor $F_{x}$. This is achieved by first extracting a set of features through the optimization process $\mathcal{O}$ on a single question-answer pair. These features are aggregated across multiple questions then concatenated to construct $F_{x}$. Repeating this process over $D_{\text{test}}$, we apply an unsupervised clustering algorithm to differentiate member documents from non-members based on their features descriptors.

Following our intuition, for each question-answer pair $(q,a)$, we fine-tune the target model parameter $\theta_{t}$  
using gradient descent to maximize the conditional probability $p_{\theta}(a|x,q)$, as defined by the objective in Equation \ref{eq:training_loss}. The optimization process always starts from the target model parameters $\theta_t$, and the learning rate $\alpha$ controls the optimization speed. During this process, we query the model at each step $s$ using $q$, tracking its prediction quality against $(q,a)$ via a utility function $\mathcal{U}$, either ACC or NLS. The optimization stops when no further improvements is observed, governed by a threshold $\tau$ or after a maximum of $S$ steps. At the end of the optimization, we evaluate the distance $\Delta$ based on Equation \ref{eq:distance}, record the number of steps taken $s$, and aggregate the utility evolution throughout the process to obtain an overall DocVQA score $u$. Collectively, these features serve as membership signals for the current $(q,a)$ pair in relation to the target document $x$.

Since each document is associated with a varying number of question-answer pairs $M$, we employ an aggregation function $\Phi$ to aggregate the features across all $M$ questions, producing in a scalar value for each feature. Optionally, we can utilize a diverse set of aggregation functions to further enrich the feature set. After aggregation, we normalize all aggregated features to ensure they are on a consistent scale. The features descriptor $F_{x}$, assigned to document $x$, is constructed by concatenating these normalized features. The assignment algorithm for each document is detailed in Algorithm \ref{alg:assignment}. Finally, we apply a clustering algorithm to the set of descriptors from documents in $D_{\text{test}}$, predicting the cluster with the larger $\Delta$ as corresponding to non-member documents.
\subsubsection{Improving Efficiency}
\begin{figure}[t]
    \centering
    \subfigure[\centering Optimize the Document]{
        \includegraphics[width=0.24\textwidth]{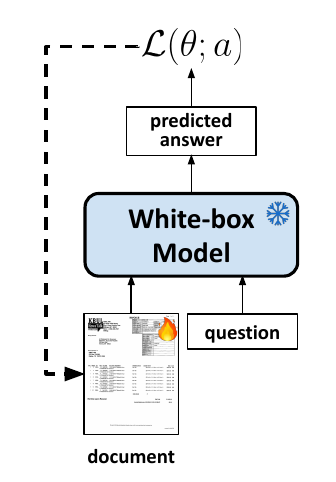}
        \label{fig:method_ig}
    }
    \subfigure[\centering Optimize One Layer]{
        \includegraphics[width=0.31\textwidth]{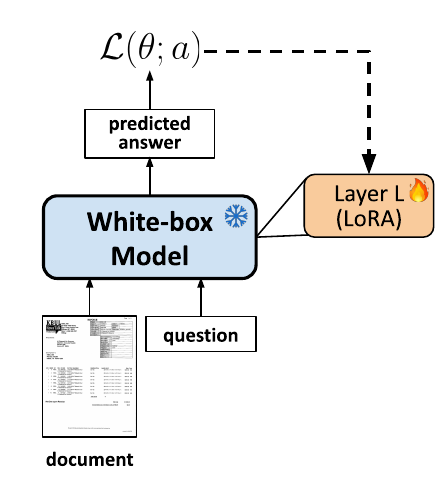}
        \label{fig:method_fl}
    }
    \subfigure[\centering Distill the Black-box]{
        \includegraphics[width=0.33\textwidth]{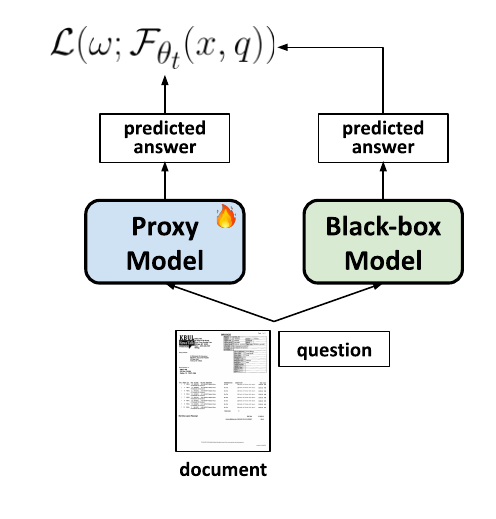}
        \label{fig:method_distill}
    }
    \caption{\textbf{Variants of our proposed \atk.} \emph{Left:} (a) (b) illustrate three attack strategies in the white-box setting: optimizing either the Document Image or a Single Layer (LoRA). Dashed arrows indicate the back-propagated gradient during optimization. \emph{Right:} We distill the black-box into a proxy model, which is then attacked using the white-box strategies.}
    \label{fig:method}
    \vspace{-0.3in}
\end{figure}

Fine-tuning $\mathcal{F}_t$ on a \textit{single} document/question-answer pair provides a mechanism to differentiate between members and non-members. However, this approach is relatively slow, given the model’s size and the complexity of data pre-processing. To improve the attack efficiency, we introduce three variants of the method, as illustrated in Figure \ref{fig:method}:

\textbf{Optimize One Layer (\atkFL)}. Instead of optimizing all parameters, we hope that gradients with respect to a single layer's parameters can provide sufficient signal for membership classification. In this variant, we select one specific layer $L$ to optimize while keeping the remaining parameters fixed. We ablate the choice of layer for this method in Appendix~\ref{sec:calibration}. In addition, we consider a variant leveraging LoRA \citep{hu2021lora}, termed \textbf{\atkFLLoRA}, where the LoRA parameters are initialized with Kaiming initialization \citep{he2015delving}. From Algorithm \ref{alg:assignment}, we replace $\theta$ to $\theta_L$ or the $\textsc{LoRA}$ parameters of the layer L, denoted as $\textsc{LoRA}(\theta_L)$, respectively.

\textbf{Optimize the Document Image (\atkIG)}.
By switching the perspective to the input space, this variant directly optimizes the pixel values of the document image $x$. The underlying intuition remains the same: training documents require less self-tuning allowing the model to converge faster to the correct answer than non-trainings. However, this assumes the target model allows differentiation of the document image pixels through its architecture. Accordingly, we replace $\theta$ with $x$ from Algorithm \ref{alg:assignment} while freezing the target model parameters $\theta$.

These variants reduce computational costs while maintaining attack performance, providing more practical options when the size of $D_\text{test}$ increases.
\vspace{-0.1in}
\subsection{Black-box \atk}
\label{sec:docmia_blackbox}
In the black-box setting, the attack model’s access is restricted to $D_{\text{test}}$ and the predicted answers. To address these limitations, we propose a distillation-based attack strategy. The key idea is to transfer knowledge about the private data $D_t$ from the black-box model $\mathcal{F}_t$ to a proxy model $\mathcal{F}_p$, parameterized by $\omega$. With full control over the proxy model, the attacks we design for the white-box setting can be fully applied to this.

Specifically, the black-box model is first employed to generate labels for each question in $D_\text{test}$, creating a query dataset $D_{\text{query}}=\{(x_i, Q_i)\}^{N_{\text{test}}}_{i=1}$ where $Q_i = \{(q_j, \mathcal{F}_{\theta_t}(x_i, q_j))\}^{M}_{j=1}$. The proxy model $\mathcal{F}_p$ is then trained on this query dataset, with the objective to maximize the likelihood of the predicted answer $p_{\omega}(\mathcal{F}_{\theta_t}(x_i, q_j)|x_i, q_j)$. In essence, the goal is to replicate the label-prediction behavior of the black-box model. By doing so, we aim to transfer the label space structure from the black-box to the proxy model, with the expectation that the membership features embedded in the black-box model will also be transferred, thus making our attack assumptions under white-box setting valid. Figure \ref{fig:method_distill} illustrates the scheme of our proposed attack.

Since our focus is on the document domain, we initialize $\mathcal{F}_p$ using a publicly available checkpoint $\omega_{\text{pt}}$, pre-trained with self-supervised learning on unlabeled document dataset $D_{\text{pt}}$, which is \textit{inaccessible} and assumed to be \textit{disjoint} from the private dataset $D_t$. This initialization equips the proxy model with a certain level of document understanding while ensuring it has no prior knowledge of the private dataset. As a result, it enables the proxy $\mathcal{F}_p$ to better mimic the prediction behavior and internal dynamics of the black-box model $\mathcal{F}_t$ after fine-tuning. 

It is important to note that, in this scenario, the adversary lacks information of the black-box training algorithm $\mathcal{T}$. This means there is no advantage in terms of model architecture or other training details when constructing the proxy model. As a result, the choice of the proxy model, optimizer, learning rate, etc., is independent of the target model. However, as we demonstrate empirically in later sections (Section \ref{sec:whitebox_results}), while there is a clear benefit when the proxy model shares the same architecture as the black-box model, our attack strategies remain effective even when using entirely different architectures. This suggests that the proposed approach is robust and can be applied without relying on specific model classes or requiring detailed knowledge of the black-box model.
\vspace{-0.1in}
\section{Experimental Setup}
\vspace{-0.1in}
\label{sec:experimental_setup}

\subsection{Target Dataset and Model}

\label{sec:target_model_dataset}
\textbf{Target Dataset}. We study two established DocVQA datasets in the literature for our analysis: \textbf{DocVQA (DVQA)} \citep{mathew2021docvqa} and \textbf{PFL-DocVQA (PFL)} \citep{tito2024privacy}. Both datasets are designed for extractive DocVQA task, where the answer is explicitly found within the document image. Each document in these datasets is accompanied by varying number of questions.

\textbf{Target Model}. We consider three state-of-the-art models which are designed for document understanding tasks: (1) \textbf{Visual T5 (VT5)} \citep{tito2024privacy} (250M parameters) follows the traditional design by utilizing OCR module to facilitate the reasoning process. It leverages the T5 model pre-trained on the C4 corpus \citep{raffel2020exploring}, along with a ViT backbone pre-trained on document data \citep{li2022dit}. (2) \textbf{Donut} \citep{Kim22Donut} (201M parameters) is one of the first end-to-end DocVQA models capable of achieving competitive performance without relying on OCR. It is pre-trained on a large collection of private synthetic documents. (3) \textbf{Pix2Struct} \citep{Lee23Pix2Struct} is another OCR-free document model with two versions: Base (282M) and Large (1.3B parameters). This model is pre-trained to perform semantic parsing on an 80M subset of the C4 corpus.

For the PFL-DocVQA dataset, we consider two targets: VT5, using the public checkpoint provided by the authors\footnote{\url{https://benchmarks.elsa-ai.eu/?ch=2}}, and Donut, which we successfully trained to achieve strong performance following the training procedure from the authors. For the DocVQA dataset, we attack four targets: VT5, Donut, and Pix2Struct (Base and Large), all with publicly available checkpoints from HuggingFace\footnote{\url{https://huggingface.co/models}} \citep{wolf-etal-2020-transformers}. In the black-box setting, we use VT5 and Donut as proxy models. To train the proxy models on the query set $D_{\text{query}}$, we initialize them with their public \textit{pre-trained} checkpoints—the same checkpoints used to fine-tune the target models on the respective target datasets, as outlined in the original papers. For more details on datasets and models, see Appendix \ref{sec:dataset} and \ref{sec:attack_implementation}.

\vspace{-0.2cm}
\subsection{Implementation}
\label{sec:implementation}
Since the optimization process involves several hyperparameters, our strategy is to tune the set of hyperparameters such that our attacks remain effective against each target model under white-box settings, which we then utilize to mount attacks on black-box models.

Assuming the knowledge of training algorithm $\mathcal{T}$ is unavailable for either white-box or black-box settings, we use Adam \citep{kingma2014adam} as the optimizer $\textsc{OPT}$ and fix this choice across all our attack experiments. We explore the impact of learning rate $\alpha$, the selected layer $L$, and we carefully tune the values of threshold $\tau$ in the ablation study (Appendix \ref{sec:calibration}). Following this, we select the optimal set of hyperparameters for each model and apply these settings in all black-box experiments. For the aggregation $\Phi$, we consider 4 aggregation functions $\{\textsc{avg}; \textsc{min}; \textsc{max}; \textsc{med}\}$ for each feature, denoted as $\Phi_\text{all}$. Throughout our experiments, we employ \textsc{KMeans} as the clustering algorithm. 

\vspace{-0.2cm}
\subsection{Evaluation Metric}
Using the official split of each target dataset, we sample 300 member documents from the training split and 300 non-member documents from the test split, resulting a total of $N_\text{test}=600$ test documents. We report \textit{Balanced Accuracy} and \textit{F1 score} as this evaluation metrics for the attack's success in the balanced setting, as in prior works \citep{salem2018ml,watson2022on,ye2022enhanced}. In addition, we evaluate our attacks using \textit{True Positive Rate (TPR) at 1\% and 3\% False Positive Rate (FPR)}, following standard practices in recent MIA literature \citep{carlini2022membership}. For all unsupervised attacks, the membership score for each document is computed as the Euclidean distance between its feature vector and the centroid of the member cluster obtained via \textsc{KMEANS}.

\subsection{Baseline}
In the \textit{black-box} setting, we evaluate three MI attacks as baselines, which only requires the predicted answer to determine membership: Score-Threshold Attack ($\textsc{Score-TA}$), Unsupervised Score-based Attack with $\textsc{avg}$ ($\textsc{Score-UA}$) \citep{tito2024privacy} and $\Phi_{\text{all}}$ ($\textsc{Score-UA}_{\text{all}}$).

For the \textit{grey-box} setting, we consider two additional baselines: Min-K\%\citep{shi2023detecting} and Min-K\%++ \citep{zhang2024min}, which assumes access to token-level probabilities of the generated answers to compute the membership score of each document.

In the \textit{white-box} setting, where loss or gradient information is accessible, we evaluate three further baselines: Loss-Threshold Attack ($\textsc{Loss-TA}$) \citep{yeom2018privacy}, Unsupervised Score+Loss Attack ($\textsc{ScoreLoss-UA}_{\text{all}}$) and Unsupervised Gradient Attack ($\textsc{Gradient-UA}$)\citep{nasr2019comprehensive}.

For detailed descriptions of these methods, we refer readers to Appendix \ref{sec:attack_baseline}.

\begin{table}[t]
\begin{center}
\begin{small}
\begin{adjustbox}{width=\textwidth}
\small
\begin{tabular}{clcccccccccccc}
\toprule
\multicolumn{2}{c}{\multirow{2}{*}{Model}} & \multicolumn{2}{c}{$\textsc{Score-TA}$} & \multicolumn{2}{c}{$\textsc{Score-UA}$} & \multicolumn{2}{c}{$\textsc{Score-UA}_{\text{all}}$} & \multicolumn{2}{c}{$\textsc{Loss-TA}$} & \multicolumn{2}{c}{$\textsc{Gradient-UA}$} & \multicolumn{2}{c}{$\textsc{ScoreLoss-UA}_{\text{all}}$}\\
\cmidrule(l){3-4}
\cmidrule(l){5-6}
\cmidrule(l){7-8}
\cmidrule(l){9-10}
\cmidrule(l){11-12}
\cmidrule(l){13-14}
&                                 & ACC & F1 & ACC & F1 & ACC & F1 & ACC & F1 & ACC & F1 & ACC & F1 \\
\midrule 
\multirow{2}{*}{\rotatebox[origin=c]{90}{PFL}}& {VT5} & \cellcolor[HTML]{C0C0C0}{$\textbf{62.33}$} & \cellcolor[HTML]{C0C0C0}{$\textbf{64.13}$} & \cellcolor[HTML]{C0C0C0}{$61.00_{0.0}$} & \cellcolor[HTML]{C0C0C0}{$68.80_{0.0}$} & \cellcolor[HTML]{C0C0C0}{$60.67_{0.0}$} & \cellcolor[HTML]{C0C0C0}{$60.67_{0.0}$} & $57.83$ & $62.81$ & $60.67_{0.0}$ & $60.67_{0.0}$ & $\textbf{60.67}_{\textbf{0.0}}$ & $\textbf{60.67}_{\textbf{0.0}}$\\

& {Donut} & \cellcolor[HTML]{C0C0C0}{$\textbf{73.33}$} & \cellcolor[HTML]{C0C0C0}{$\textbf{75.14}$} & \cellcolor[HTML]{C0C0C0}{$68.33_{0.0}$} & \cellcolor[HTML]{C0C0C0}{$75.77_{0.0}$} & \cellcolor[HTML]{C0C0C0}{$71.17_{0.67}$} & \cellcolor[HTML]{C0C0C0}{$71.33_{3.32}$} & $\textbf{73.67}$ & $\textbf{78.99}$ & $70.67_{0.0}$ & $69.55_{0.0}$ & $70.83_{0.0}$ & $69.67_{0.0}$
\\
\midrule

\multirow{3}{*}{\rotatebox[origin=c]{90}{DVQA}} & {VT5} & \cellcolor[HTML]{C0C0C0}{$\textbf{75.67}$} & \cellcolor[HTML]{C0C0C0}{$\textbf{75.75}$} & \cellcolor[HTML]{C0C0C0}{$72.17_{0.0}$} & \cellcolor[HTML]{C0C0C0}{$76.18_{0.0}$} & \cellcolor[HTML]{C0C0C0}{$75.17_{0.0}$} & \cellcolor[HTML]{C0C0C0}{$75.13_{0.0}$} & $73.67$ & $77.99$ & $71.17_{0.0}$ & $67.54_{0.0}$ & $\textbf{75.50}_{\textbf{0.0}}$ & $\textbf{76.02}_{\textbf{0.0}}$
\\

& {Donut} & \cellcolor[HTML]{C0C0C0}{$79.67$} & \cellcolor[HTML]{C0C0C0}{$79.53$} & \cellcolor[HTML]{C0C0C0}{$75.97_{0.07}$} & \cellcolor[HTML]{C0C0C0}{$79.57_{0.07}$} & \cellcolor[HTML]{C0C0C0}{$\textbf{80.50}_{\textbf{0.0}}$} & \cellcolor[HTML]{C0C0C0}{$\textbf{81.10}_{\textbf{0.0}}$} & $51.83$ & $53.46$ & $77.17_{0.0}$ & $75.92_{0.0}$ & $\textbf{80.50}_{\textbf{0.0}}$ & $\textbf{81.10}_{\textbf{0.0}}$
\\

& {Pix2Struct-B} & \cellcolor[HTML]{C0C0C0}{$67.33$} & \cellcolor[HTML]{C0C0C0}{$67.97$} & \cellcolor[HTML]{C0C0C0}{$\textbf{68.17}_{\textbf{0.0}}$} & \cellcolor[HTML]{C0C0C0}{$\textbf{71.36}_{\textbf{0.0}}$} & \cellcolor[HTML]{C0C0C0}{$69.13_{0.07}$} & \cellcolor[HTML]{C0C0C0}{$67.67_{0.09}$} & $59.33$ & $64.63$ & $66.0_{0.0}$ & $68.32_{0.0}$ & $\textbf{69.00}_{\textbf{0.0}}$ & $\textbf{67.48}_{\textbf{0.0}}$
\\
\bottomrule
\end{tabular}
\end{adjustbox}
\end{small}
\end{center}
\caption{\textbf{Results from Baseline Attacks.} \textcolor{gray}{\textbf{Gray}} color indicate attacks conducted in the black-box setting. All results are reported based on five random seeds for \textsc{KMeans}. The methods with the best \textit{average} performance across the two metrics are highlighted in \textbf{bold}.}
\vspace{-0.2in}
\label{tab:baseline_results}
\end{table}

\vspace{-0.1in}
\section{Evaluation}
\label{sec:evaluation}
\vspace{-0.1in}
\subsection{White-box Setting}
\label{sec:whitebox_results}

\textbf{Baseline Performance Evaluation.}
Table \ref{tab:baseline_results} (\textit{right}) shows the performance of baseline attacks in the white-box setting. $\textsc{Loss-TA}$, akin to the thresholding loss attack in \citep{yeom2018privacy}, performs poorly on complex DocVQA models, achieving under 60\% accuracy for most targets. In contrast, $\textsc{ScoreLoss-UA}_{\text{all}}$, which combines utility scores and loss features, achieves stronger results: 81\% F1 on Donut, 75\% on VT5, and 69\% on Pix2Struct on DocVQA dataset. However, it underperforms $\textsc{Loss-TA}$ on PFL-DocVQA, with a 3\% drop in Accuracy and 8\% in F1, likely due to high loss variance in this dataset. $\textsc{Gradient-UA}$, which incorporates one-step gradient information, matches the performance of score-based attacks, suggesting that the gradient serves as a useful signal for membership inference. However, none of the baselines generalizes well across all target models.

\begin{figure}[t]
    \centering
    \begin{minipage}{0.48\textwidth}
        \centering
        \resizebox{0.8\linewidth}{!}{
            \includegraphics[width=1.0\linewidth]{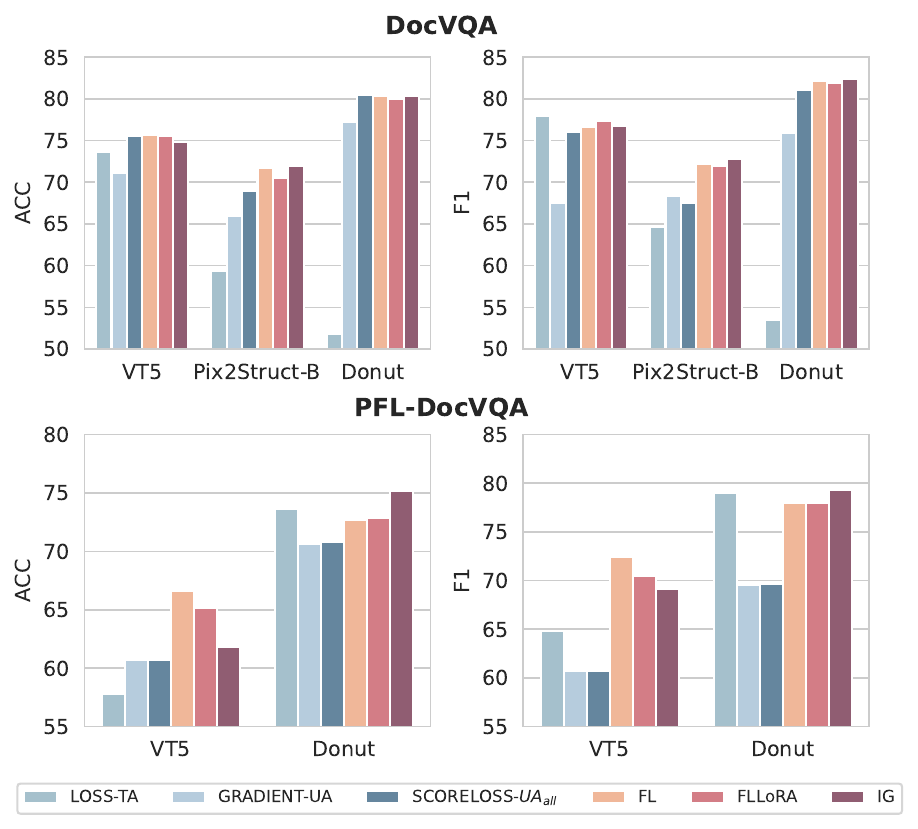}
        }
        \caption{\textbf{White-box Setting: Our proposed attacks consistently achieve high performance}, generally outperforming the considered baselines.}
        \label{fig:whitebox_results}
        \end{minipage}
        \hfill
    \begin{minipage}{0.51\textwidth}
        \centering
        \resizebox{\linewidth}{!}{
            \begin{tabular}{lccccc}
                \toprule
                 & \multicolumn{3}{c}{\textbf{DVQA}} & \multicolumn{2}{c}{\textbf{PFL}}  \\
                \cmidrule(l){2-4}
                \cmidrule(l){5-6}
                 & {VT5} & {Donut} & {Pix2Struct-B} & {VT5} & {Donut} \\
                \midrule
                $\textsc{Loss-TA}$ & \textbf{14.00} & 7.67 & 5.33 & 3.00 & \textbf{14.67} \\
                $\textsc{Gradient-UA}$ & 9.33 & 6.00 & 5.00 & 3.00 & 8.33 \\
                $\textsc{ScoreLoss-UA}_{\text{all}}$ & 4.67 & 8.67 & 6.67 & 4.00 & 6.33 \\
                \midrule
                Min-K\% & 10.67 & 1.33 & 5.33 & 5.67 & 0.00 \\
                Min-K\%++ & 7.00 & 9.33 & 10.33 & 8.00 & 2.00 \\
                \midrule
                FL & 5.67 & \textbf{10.67} & \textbf{11.00}& \textbf{8.67} & 7.00 \\
                FLLoRA & 11.33 & 5.33 & 6.33 & 3.33 & 10.00 \\
                IG & 5.67 & 8.00 & 10.33 & 2.33  & 11.00 \\
                \bottomrule
            \end{tabular}
        }
        \captionof{table}{\textbf{White-box Setting: TPR at 3\% FPR}. Comparison across all white-box methods, with the best-performing method for each metric highlighted in \textbf{bold}. We refer the readers to Appendix \ref{sec:tpr_at_fix_fpr} for the complete results.}
        \label{tab:tpr@fpr_main_whitebox}
        \end{minipage}
    \vskip -0.2in
\end{figure}

\textbf{Our Proposed Attacks outperform the Baselines.} We evaluate our proposed attacks—\atkFL, \atkFLLoRA, and \atkIG—in the white-box setting across target models. As shown in Figure \ref{fig:whitebox_results}, our methods consistently achieve high performance, indicating that \textit{optimization-based features generalize well} across various models.
Compared to all baselines, our attacks achieve either the best or near-best performance on both target datasets, with notable F1 scores of 72\% against VT5 and Pix2Struct, and 82.5\% against Donut. Against $\textsc{Gradient-UA}$, our optimization-based features yield up to a 10\% improvement in F1 on Donut, indicating that \textit{single-step gradients are insufficient} for reliable membership inference.
From Table \ref{tab:tpr@fpr_main_whitebox}, our attacks consistently perform well in the low-FPR regime, often surpassing or matching the strongest baselines. For instance, \atkFL achieves a TPR of 8.67\% at 3\% FPR against VT5 on PFL, despite minimal overfitting, and a TPR of 11.00\% at the same FPR against Pix2Struct-B on DocVQA. Additionally, our methods outperform both Min-K\% and Min-K\%++ across all target models, underscoring their effectiveness, particularly for DocMIA setting.
These results highlight the privacy risks posed by optimization-based features in membership inference. For full results and in-depth analysis, please refer to  Appendix \ref{sec:tpr_at_fix_fpr} and \ref{sec:more_analysis}.

\textbf{Why are Our Attacks more effective?}
We evaluate the effectiveness of our optimization-based features compared to traditional metrics such as loss or single-gradient norm.

The Loss-based attack $\textsc{LOSS-TA}$ assumes that member documents exhibit lower loss values than non-member documents after training the target model $\mathcal{F}_{t}$. While this approach leverages the generalization gap, it proves too simplistic for large-scale models that are trained with complex training process to minimize overfitting. The generalization capability of these models, especially in DocVQA tasks, often reduces the sensitivity of the loss as a membership indicator. Our attacks, on the other hand, leverage the optimization landscape with respect to the model parameters, conditioned on each question-answer pair. We hypothesize that the \textit{distance} resulting from parameter optimization of pairs from a member document will be smaller compared to those for a non-member document, as depicted in Figure \ref{fig:intuition}. This fine-grained signal, which reflects the model's internal response to optimization, offers a more discriminative feature for identifying membership.

As illustrated in Appendix Figure \ref{fig:whitebox_distance}, our \textit{distance} feature, computed from the optimization process, provides a better separation between members and non-members compared to loss-based methods (Figure \ref{fig:whitebox_loss}  (\textit{top})). The t-SNE visualization \citep{JMLR:v9:vandermaaten08a} from Figure \ref{fig:whitebox_loss} (\textit{bottom}) further demonstrates that features derived from our attacks yield a more distinct clustering of member and non-member documents in high-dimensional space for all target models, underscoring its efficacy as a membership indicator, therefore outperforms the loss-based approach.
\vspace{-0.1in}
\subsection{Black-box Setting}
\label{sec:blackbox_results}
\begin{figure}[h]
    \centering
    \subfigure[\centering \textbf{White-box}]{
        \includegraphics[width=0.48\textwidth]{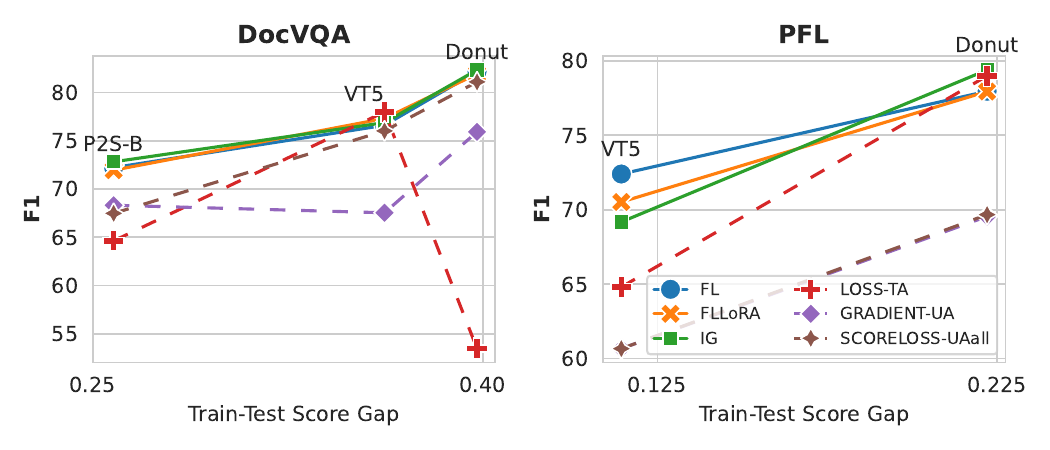}
    }
    \subfigure[\centering \textbf{Black-box}]{
        \includegraphics[width=0.48\textwidth]{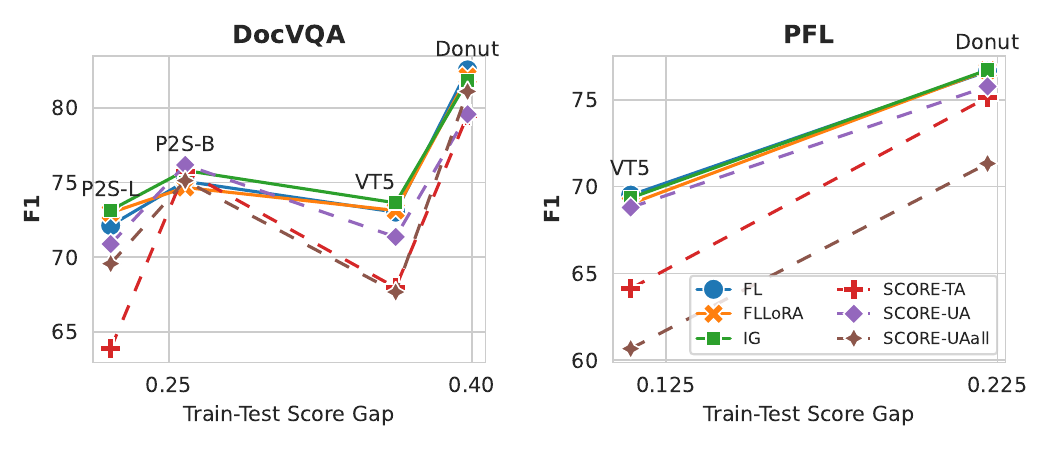}
    }
\caption{\textbf{MI performance versus the Train-Test gap.} The target models exhibit varying Train-Test gaps, measured by the difference in DocVQA scores between member/non-member documents. Our attacks remain effective even when the gap is small, with performance improving as the gap increases across most target models and datasets. In contrast, baseline methods show more variable performance under these conditions.}
\vspace{-0.1in}
\label{fig:F1_scoregap}
\end{figure}

\textbf{Baseline Performance Evaluation.} Table \ref{tab:baseline_results} (\textit{left}) presents the results of our black-box baseline attacks, all of which rely on the DocVQA score as the only source of information in this setting. Similar to the loss metric, the score metric is directly correlated with the generalization gap, making attacks more effective when there is a higher degree of overfitting. This trend is illustrated in Figure \ref{fig:F1_scoregap}, where we observe strong MI performance, particularly for the Donut model with 75\% in PFL and 79\% F1 score in DocVQA. Meanwhile, both $\textsc{Score-UA}$-based baselines show comparable performance, especially effective against models trained on DocVQA. Overall, no single method emerges as the clear winner across all target models.

\textbf{Attacking via Proxy Model.} Table \ref{tab:blackbox_results} and Table \ref{tab:tpr@fpr_main_blackbox} present the key results of our proposed black-box attacks using two proxy models, VT5 and Donut. Several important observations:

First, we observe a clear advantage of attacking the proxy models distilled with our proposed techniques. Across a wide range of black-box architectures trained on both target datasets, attacks leveraging the proxy models outperform the black-box baselines in most cases, demonstrating better MI performance. This suggests that, even without knowledge of the black-box model architecture, \textit{one chosen proxy model still effectively distills certain behaviors} from the black-box models which are membership-indicative, enabling our attacks to infer membership with high accuracy.

When the black-box model architecture \textit{matches} that of the proxy, we consistently observe improvements in MI performance, especially when targeting the PFL-DocVQA dataset. Among the target, \textit{Pix2Struct is the most vulnerable} (both Base and Large). Both VT5 and Donut proxies gains of +3.04\% in Accuracy and +4.88\% in F1 score over the best baseline, even against the Pix2Struct-L model, which exhibits strong generalization and a minimal Train-Test gap (Figure \ref{fig:F1_scoregap}).
Furthermore, proxy VT5 can achieve TPRs of 23.00\% and 16.67\% against Donut and Pix2Struct-B, respectively, at 3\% FPR on DocVQA.
We also provide an analysis of the proxy model in Appendix \ref{sec:impact_proxy_model}.

These results suggest that privacy vulnerabilities can be exploited using simple distillation-based strategies applied to the model's output space.

\begin{table}[t]
    \centering
    \begin{minipage}{0.6\textwidth}
        \centering
        \resizebox{\linewidth}{!}{
            \begin{tabular}{clcccccc}
            \toprule
            \multicolumn{2}{c}{Proxy Model} & \multicolumn{6}{c}{VT5}\\
            
            \midrule
            \multicolumn{2}{c}{\multirow{2}{*}{\textbf{Black-box}}} & \multicolumn{2}{c}{FL} & \multicolumn{2}{c}{FLLoRA} & \multicolumn{2}{c}{IG}\\
            \cmidrule(l){3-4}
            \cmidrule(l){5-6}
            \cmidrule(l){7-8}
            &                                 & ACC & F1 & ACC & F1 & ACC & F1\\
            \midrule 
            \multirow{2}{*}{\rotatebox[origin=c]{90}{\textbf{PFL}}}& \textbf{VT5} & $63.33_{0.0}\textcolor{red}{(+2.33)}$ & $69.51_{0.03}\textcolor{red}{(+0.71)}$ & $63.33_{0.0}\textcolor{red}{(+2.33)}$ & $69.01_{0.0}\textcolor{red}{(+0.21)}$ & $62.00_{0.16}\textcolor{red}{(+1)}$ & $69.35_{0.2}\textcolor{red}{(+0.55)}$\\
            
            & \textbf{Donut} & $70.83_{0.0}\textcolor[HTML]{0b5394}{(-0.34)}$ & $76.64_{0.0}\textcolor{red}{(+0.87)}$ & $70.83_{0.0}\textcolor[HTML]{0b5394}{(-0.34)}$ & $76.70_{0.0}\textcolor{red}{(+0.93)}$ & $70.67_{0.0}\textcolor[HTML]{0b5394}{(-0.5)}$ & $76.72_{0.0}\textcolor{red}{(+0.95)}$\\
            \midrule
            
            \multirow{4}{*}{\rotatebox[origin=c]{90}{\textbf{DVQA}}}& \textbf{VT5} & $74.33_{0.01}\textcolor[HTML]{0b5394}{(-0.84)}$ & $75.08_{0.0}\textcolor[HTML]{0b5394}{(-1.1)}$ & $74.33_{0.0}\textcolor[HTML]{0b5394}{(-0.84)}$ & $74.67_{0.0}\textcolor[HTML]{0b5394}{(-1.51)}$ & $73.83_{0.08}\textcolor[HTML]{0b5394}{(-1.34)}$ & $75.81_{0.0}\textcolor[HTML]{0b5394}{(-0.37)}$\\
            
            & \textbf{Donut} & $81.67_{0.0}\textcolor{red}{(+1.17)}$ & $82.54_{0.0}\textcolor{red}{(+1.44)}$ & $81.17_{0.0}\textcolor{red}{(+0.67})$ & $82.09_{0.0}\textcolor{red}{(+0.99)}$ & $80.17_{0.0}\textcolor[HTML]{0b5394}{(-0.33)}$ & $81.89_{0.0}\textcolor{red}{(+0.79)}$\\
            
            & \textbf{Pix2Struct-B} & $70.17_{0.0}\textcolor{red}{(+1.04)}$ & $69.71_{0.0}\textcolor[HTML]{0b5394}{(-1.65})$ & $70.27_{0.23}\textcolor{red}{(+1.14)}$ & $70.85_{0.07}\textcolor[HTML]{0b5394}{(-0.51)}$ & $71.17_{0.0}\textcolor{red}{(+2.04)}$ & $72.14_{0.0}\textcolor{red}{(+0.78)}$\\
            
            \cmidrule{2-8}
            & \textbf{Pix2Struct-L} & $71.67_{0.01}\textcolor{red}{(+0.84)}$ & $72.13_{0.0}\textcolor{red}{(+1.30)}$ & $70.17_{0.0}\textcolor[HTML]{0b5394}{(-0.66)}$ & $71.27_{0.0}\textcolor{red}{(+0.44)}$ & $71.00_{0.05}\textcolor{red}{(+0.17)}$ & $73.15_{0.0}\textcolor{red}{(+2.32)}$\\
            
            \midrule
            \multicolumn{2}{c}{Proxy Model} & \multicolumn{6}{c}{Donut}\\ 
            
            \midrule
            \multirow{2}{*}{\rotatebox[origin=c]{90}{\textbf{PFL}}}& \textbf{VT5} & $61.73_{0.08}\textcolor{red}{(+0.73)}$ & $64.04_{0.10}\textcolor[HTML]{0b5394}{(-4.76)}$ & $61.67_{0.08}\textcolor{red}{(+0.67)}$ & $63.49_{0.0}\textcolor[HTML]{0b5394}{(-5.31)}$ & $55.17_{0.17}\textcolor[HTML]{0b5394}{(-5.83)}$ & $57.37_{0.3}\textcolor[HTML]{0b5394}{(-11.43)}$\\
            
            & \textbf{Donut} & $72.17_{0.0}\textcolor{red}{(+2.33)}$ & $76.24_{0.0}\textcolor[HTML]{0b5394}{(-0.19)}$ & $72.67_{0.0}\textcolor{red}{(+1.5)}$ & $77.47_{0.0}\textcolor{red}{(+1.7)}$ & $74.50_{0.0}\textcolor{red}{(+3.33)}$ & $76.43_{0.0}\textcolor{red}{(+0.66)}$\\
            
            \midrule
            \multirow{4}{*}{\rotatebox[origin=c]{90}{\textbf{DVQA}}}& \textbf{VT5} & $73.50_{0.0}\textcolor[HTML]{0b5394}{(-4.34)}$ & $75.58_{0.0}\textcolor[HTML]{0b5394}{(-4.36)}$ & $74.17_{0.0}\textcolor[HTML]{0b5394}{(-1)}$ & $76.04_{0.0}\textcolor[HTML]{0b5394}{(-0.14)}$ & $74.0_{0.0}\textcolor[HTML]{0b5394}{(-1.17)}$ & $75.93_{0.01}\textcolor[HTML]{0b5394}{(-0.25)}$\\
            
            & \textbf{Donut} & $79.50_{0.0}\textcolor[HTML]{0b5394}{(-1)}$ & $81.50_{0.0}\textcolor{red}{(+0.4)}$ & $80.0_{0.0}\textcolor[HTML]{0b5394}{(-0.5)}$ & $81.82_{0.0}\textcolor{red}{(+0.72)}$ & $80.27_{0.0}\textcolor[HTML]{0b5394}{(-0.23)}$ & $81.96_{0.0}\textcolor{red}{(+0.86)}$\\
            
            & \textbf{Pix2Struct-B} & $70.83_{0.0}\textcolor{red}{(+3.04)}$ & $71.82_{0.0}\textcolor{red}{(+4.88)}$ & $70.83_{0.06}\textcolor{red}{(+1.70)}$ & $71.73_{0.14}\textcolor{red}{(+0.37)}$ & $71.0_{0.01}\textcolor{red}{(+1.87)}$ & $71.94_{0.0}\textcolor{red}{(+0.58)}$\\
            
            \cmidrule{2-8}
            & \textbf{Pix2Struct-L} & $70.83_{0.0}{(0)}$ & $72.95_{0.0}\textcolor{red}{(+2.12)}$ & $71.0_{0.0}\textcolor{red}{(+0.17)}$ & $72.98_{0.0}\textcolor{red}{(+2.15)}$ & $71.0_{0.03}\textcolor{red}{(+0.17)}$ & $72.81_{0.01}\textcolor{red}{(+1.98)}$\\
            \bottomrule
            \end{tabular}
        }
        \caption{\textbf{Black-Box Setting: Main Results of Black-Box \atk using Donut and VT5 as proxy models}. The checkpoints for the \textbf{black-box models} are trained on the respective datasets. Values in parentheses indicate the improvement (\textcolor{red}{positive}/\textcolor[HTML]{0b5394}{negative}) compared to the \textit{best} number from $\textsc{Score-UA}$-based baselines. Results are reported over five random seeds.}
        \vspace{-0.2in}
        \label{tab:blackbox_results}
    \end{minipage}
    \hfill
    \begin{minipage}{0.39\textwidth}
        \centering
        \resizebox{\linewidth}{!}{
            \begin{tabular}[]{clcccccc}
            \toprule
             &  & \multicolumn{4}{c}{\textbf{DVQA}} & \multicolumn{2}{c}{\textbf{PFL}} \\
            \cmidrule(l){3-6}
            \cmidrule(l){7-8}
             &  & {\textbf{VT5}} & {\textbf{Donut}} & {\textbf{P2S-B}} & {\textbf{P2S-L}} & {\textbf{VT5}} & {\textbf{Donut}} \\
            \midrule
            & {$\textsc{Score-TA}$} & 9.33  & 11.00 & 8.00  & \textbf{9.00} & 5.00  & 2.67 \\
            & {$\textsc{Score-UA}$} & 7.67 & 15.67 & 6.33  & 6.67  & 3.33  & 3.33 \\
            & {$\textsc{Score-UA}_{\text{all}}$} & 9.33  & 11.00  & 8.00  & 9.00 & 5.00  & 2.67 \\
            \midrule
            \multirow{3}{*}{VT5} & FL & \textbf{12.33} & \textbf{23.00} & \textbf{16.67}  & 5.33 & 2.00 &  \textbf{8.00} \\
            & FLLoRA & {11.33} & 16.33  & 9.33  & 4.67  & 3.33 & 2.00 \\
            & IG & 8.33  & 7.00 & 7.67  & 7.00 & 3.67 & 6.67 \\
            \cmidrule(l){1-8}
            \multirow{3}{*}{Donut} & FL  & 6.33  & 4.00  & 4.67 & 7.33  & 1.33  & 4.00 \\
            & FLLoRA & 6.33  & 5.00  & 6.33  & 8.00  & 5.33 & 5.33 \\
            & IG  & 5.00 & 11.00  & 9.33 &  6.33  & \textbf{6.33}& 4.33 \\
            \bottomrule
            \end{tabular}
        }
        \caption{\textbf{Black-box Setting: TPR at 3\% FPR using Donut and VT5 as proxy models}. Comparison across all black-box methods, with the best-performing method for each metric highlighted in \textbf{bold}. The complete results can be found in the Appendix \ref{sec:tpr_at_fix_fpr}.}
        \label{tab:tpr@fpr_main_blackbox}    
    \end{minipage}
\end{table}

\vspace{-0.1in}
\section{Conclusion}
In this paper, we introduce the first document-level membership inference attacks for DocVQA models, highlighting privacy risks in multi-modal settings. By leveraging model optimization techniques, we extract discriminative features that address challenges posed by multi-modal data, repeated document occurrences in training, and auto-regressive outputs. This enables us to propose novel, auxiliary data-free attacks for both white-box and black-box scenarios. Our methods, evaluated across multiple datasets and models, outperform existing membership inference baselines, emphasizing the privacy vulnerabilities in DocVQA models and the urgent need for stronger privacy safeguards.


\section{Ethics Statement}
Our research introduces two novel membership inference attacks on DocVQA models, designed to evaluate the privacy risks inherent in such systems. While our methodology exposes vulnerabilities that could potentially be exploited for malicious purposes, the primary objective of this work is to raise awareness about privacy issues in AI systems, specifically in the context of DocVQA models, and to encourage the development of more privacy-preserving technologies.

\subsubsection*{Acknowledgment}
This work has been funded by the European Lighthouse on Safe and Secure AI (ELSA) from the European Union’s Horizon Europe programme under grant agreement No 101070617. Views and opinions expressed are however those of the authors only and do not necessarily reflect those of the European Union or European Commission. Neither the European Union nor the European Commission can be held responsible for them. Khanh Nguyen and Dimosthenis Karatzas have been supported by the Consolidated Research Group 2021 SGR 01559 from the Research and University Department of the Catalan Government, and by project PID2023-146426NB-100 funded by MCIU/AEI/10.13039/501100011033 and FSE+.


\bibliography{main}
\bibliographystyle{iclr2025_conference}

\appendix

\newpage

\startcontents[appendices]
\printcontents[appendices]{l}{1}{\section*{\textbf{Appendix}}\setcounter{tocdepth}{4}}

\vspace{1cm}

\section{DocVQA Datasets}
\label{sec:dataset}

\textbf{DocVQA} \citep{mathew2021docvqa} This dataset contains high-quality human annotations and is widely used as a benchmark for document understanding. It comprises of real-world administrative documents across a diverse range of types, including letters, invoices, and financial reports.

\textbf{PFL-DocVQA} \citep{tito2024privacy} A large-scale dataset of real business invoices, often containing privacy-sensitive information such as payment amounts, tax numbers, and bank account details. This dataset is specifically designed for DocVQA tasks in a federated learning and differential privacy setup, supporting different levels of privacy granularity. The dataset is accompanied by a variant of MI attacks, where the goal is to infer the membership of the invoice's owner (i.e., the provider) from a set of their invoices that were not used during training.

\begin{figure}
    \centering
    \subfigure[\centering \textbf{DocVQA}]{
    \includegraphics[height=3cm,width=0.45\textwidth]{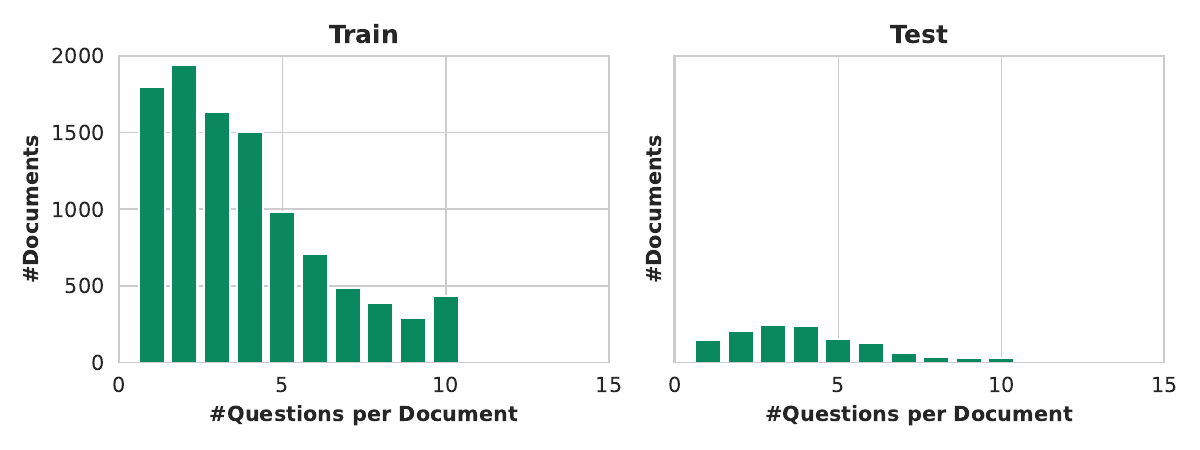}
        \label{fig:docvqa_dist}
    }
    \subfigure[\centering \textbf{PFL-DocVQA}]{
    \includegraphics[height=3cm,width=0.45\textwidth]{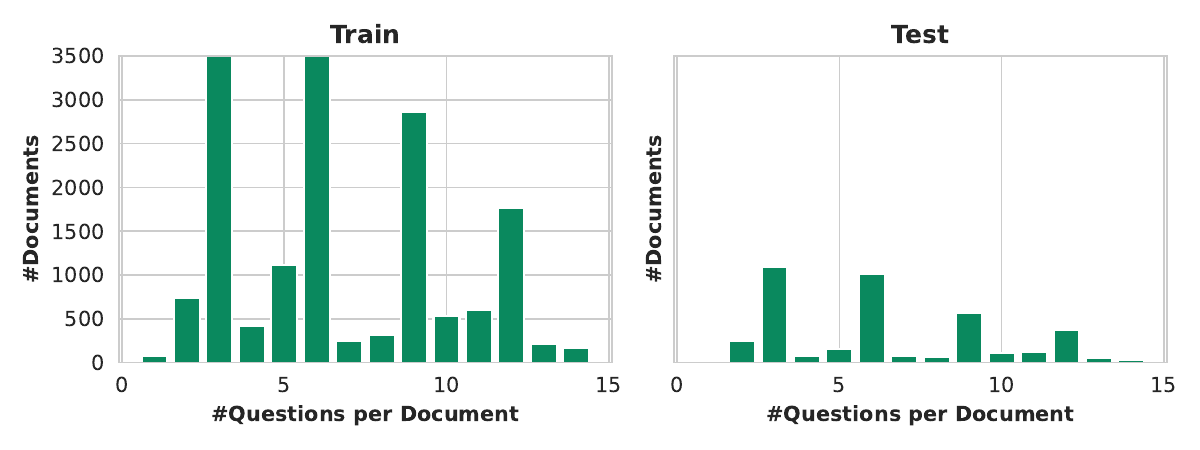}
        \label{fig:pfl_dist}
    }
\caption{\textbf{The distribution of number per-document questions} from PFL and DocVQA dataset.}
\label{fig:questions_per_document}
\end{figure}

\begin{table}[H]
\begin{center}
\begin{small}
\small
\begin{tabular}{lcccc}
\toprule
 & \multicolumn{2}{c}{DocVQA} & \multicolumn{2}{c}{PFL-DocVQA} \\
\cmidrule(l){2-3}
\cmidrule(l){4-5}
Split & Num. Docs & Num. Questions & Num. Docs & Num. Questions \\
\midrule 
Train & 69894 & 221316 & 10194 & 39463\\
Val & 9150 & 30491 & 1286 & 5349\\
Test & 13463 & 43591 & 1287 & 5188\\
\bottomrule
\end{tabular}
\end{small}
\end{center}
\vspace{-0.1in}
\caption{\textbf{Statistics} from PFL and DocVQA dataset.}
\label{tab:dataset_stats}
\vspace{-0.1in}
\end{table}
In Table \ref{tab:dataset_stats}, we present statistics for both the DocVQA and PFL-DocVQA datasets. Additionally, Figure \ref{fig:questions_per_document} shows the distribution of questions per document: (1) while a small subset of documents have more than 10 questions, most contain fewer, and (2) a fraction of documents have only a single question. These trends hold across both datasets.


\section{Baselines}
\label{sec:attack_baseline}
For the \textit{black-box} setting, we evaluate three MI attacks as baselines, which only requires generated text to infer the membership of the target document:

\textbf{Score-Threshold Attack $\textsc{(Score-TA)}$} 
assumes that training documents should achieve higher scores than non-training ones. This attack, adapted from \citet{yeom2018privacy}, evaluates the prediction $\hat{a}$ for each question $q$ using the utility function $\mathcal{U}$ and computes the average score $\bar{u}$. A document is then predicted as a member $\bar{u} \geq \kappa$, and non-member otherwise. The threshold $\kappa$ is set as the average value of $\bar{u}$ across $D_\text{test}$.

\textbf{Unsupervised Score-based Attack $\textsc{(Score-UA)}$} \citep{tito2024privacy}. This attack applies an unsupervised clustering algorithm over the set of average score $\bar{u}$ from test documents in $D_\text{test}$, documents within the cluster with higher average score are predicted as members.

\textbf{Unsupervised Score-based Attack - An Extension ($\textsc{Score-UA}_{\text{all}}$)}. This attack extends $\textsc{Score-UA}$ by considering multiple aggregation functions $\Phi_\text{all}$ to form the feature vector.

For the \textit{grey-box} setting, we consider two additional baselines which assumes access to token-level probabilities of the generated answers $a$ to compute the membership score of each document:

\textbf{Min-K\%}\citep{shi2023detecting} computes the average log probability of the lowest-K\% answer tokens as the membership score: $\text{Min-K\%} = \frac{1}{|\text{Min-K\%}(a)|}\Sigma_{a_i\in\text{Min-K}(a)} \log p(a_i| a_{<i})$. Intuitively, training documents are less likely to contain low-probability answer tokens, resulting in higher  scores.

\textbf{Min-K\%++}\citep{zhang2024min} also averages scores from the lowest-K\% probability tokens but assumes that tokens in the predicted answers for training documents have high probabilities or often form the mode of the conditional distribution. Thus, for each token, the score is computed as: $\text{Min-K\%++}(a_{<i}, a_i) = \frac{\log p(a_i| a_{<i}) - \mu_{a_{<t}}}{\sigma_{a_{<t}}}$ with $\mu_{a_{<t}}$ and $\sigma_{a_{<t}}$ are the expectation and standard deviation of $p(a_i| a_{<i})$ respectively.

We adapt these baselines to DocMIA by using an \textsc{AVG} aggregation function to combine scores across question-answer pairs within a document. We evaluated $K\in [0.6, 0.7, 0.8, 0.9, 1.0]$, which correspond to corresponds to 60\% to 100\% the length of the answer and reported the best result.

In the \textit{white-box} setting, where loss information is available, we consider three additional baselines:

\textbf{Loss-Threshold Attack ($\textsc{Loss-TA}$)} \citep{yeom2018privacy}
Similar to $\textsc{Score-TA}$, this attack computes the average loss $\bar{l}= \frac{1}{M}\Sigma^{M}_{i} \mathcal{L}(\mathcal{F}(x, q_i))$. A document is predicted as a member if $\bar{l} \leq \kappa$ and otherwise non-member, where $\kappa$ is selected as the average value of $\bar{l}$ across $D_\text{test}$.

\textbf{Unsupervised One-step Gradient Attack ($\textsc{Gradient-UA}$)} Inspired from \citet{nasr2019comprehensive}, this attack utilizes the average norm of the gradient of the loss $\nabla_{\theta}\mathcal{L}$ from a single optimization step. It also incorporates the average score $\bar{u}$, both aggregated with $\Phi_\text{all}$ as the features to perform clustering.

\textbf{Unsupervised Score+Loss Attack ($\textsc{ScoreLoss-UA}_{\text{all}}$)}
This attack extends $\textsc{ScoreUA}_{\text{all}}$, combining the average loss $\bar{l}$ with the average utility score $\bar{u}$, then aggregating with $\Phi_\text{all}$.

\section{Ablation Study}
\label{sec:calibration}
In this section, we provide a detailed analysis of the hyperparameter tuning process for \atk in the white-box setting, targeting all the considered models. Given the high computational cost due to the numerous factors involved, we focus on the key parameters that may potentially affect the attack performance. Our intuition behind this tuning process is that: achieving a reliable estimate of the distance $\Delta$ requires the optimization process to converge effectively, which in turn correlates with higher attack accuracy. Thus in all of our experiments, to increase the likelihood of convergence, we set the maximum number of optimization steps to $S = 200$. We fix the maximum number of questions per document $M$ to 10.

\textbf{Learning Rate $\alpha$}. We first study the effect of $\alpha$, which controls the speed of the optimization process in our attacks. This threshold $\tau$ is empirically set to be the average loss change observed when performing one optimization step after reaching the correct answer. Only the distance $\Delta$ and the number of steps $s$ are used as the features. For FL and FLLoRA attacks, we perform a hyperparameter search over a grid of learning rates, $\alpha \in \{10^{-4}, 0.001, 0.01, 0.1, 0.5, 1.0\}$, and $\alpha \in \{0.001, 0.01, 0.1, 0.5, 1.0, 5.0, 10.0, 20.0\}$ for the IG attacks. For FL and FLLoRA, we specifically tune the embedding projection layer, which projects the final hidden states into the vocabulary space, a common design choice across all the target models considered.

\begin{figure}[t]
    \centering
    \subfigure[\centering \textbf{Learning Rate $\alpha$}]{
        \includegraphics[height=3cm,width=0.45\textwidth]{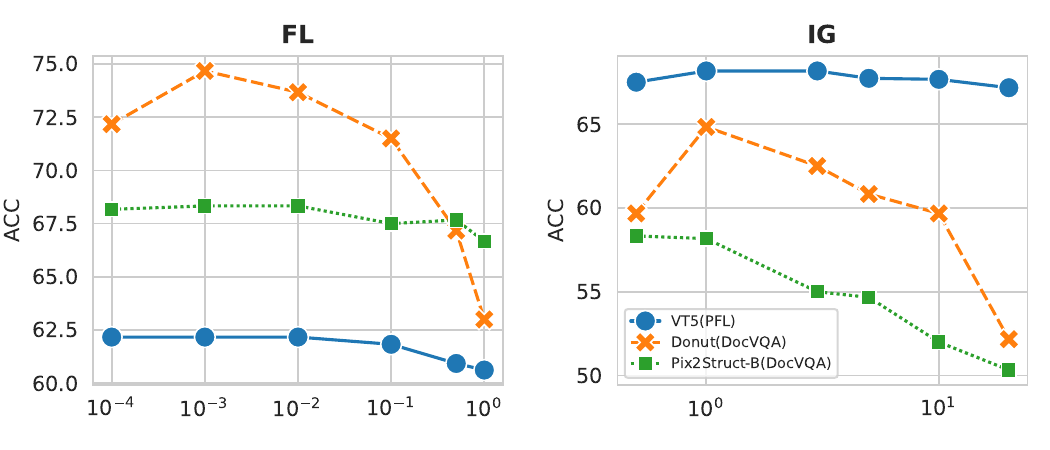}
        \label{fig:ablate_alpha}
    }
    \subfigure[\centering \textbf{Threshold $\tau$}]{
        \includegraphics[height=3cm,width=0.45\textwidth]{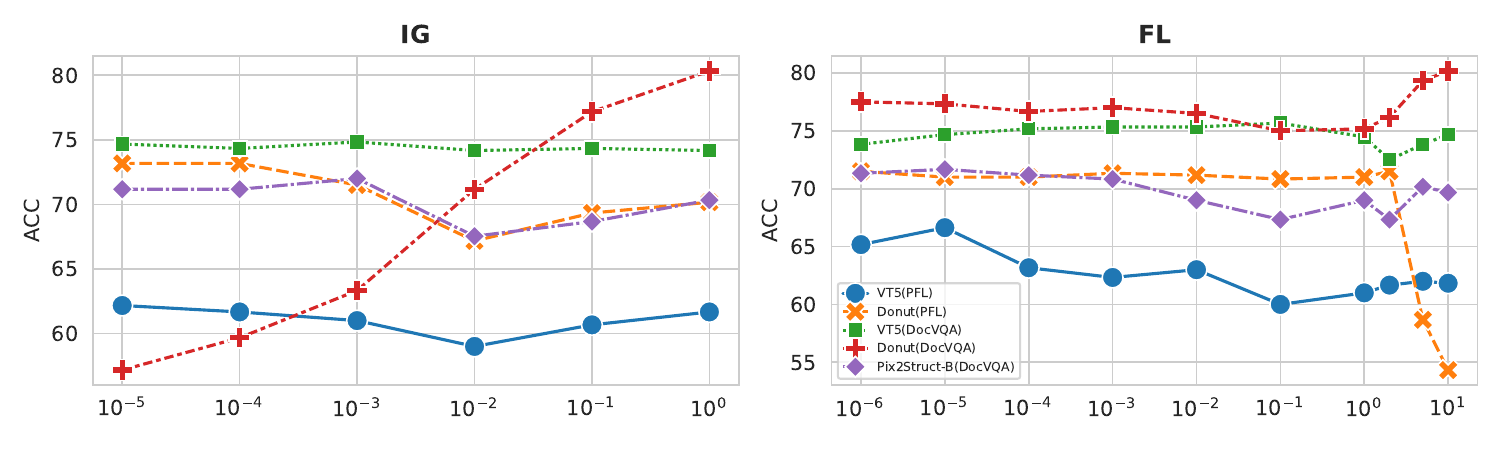}
        \label{fig:ablate_tau}
    }
\caption{\textbf{Ablation Study on Learning Rate $\alpha$ and Threshold $\tau$.} The best value for each model across all datasets is used as the hyperparameters in our black-box attacks.}
\label{fig:ablate_alpha_tau}
\end{figure}

As shown in Figure \ref{fig:ablate_alpha}, setting a high learning rate can cause the optimization process to overshoot, while lower values lead to a more stable but slower convergence. We find that a learning rate of $\alpha = 10^{-3}$ consistently delivers the best attack performance across most of the settings.

\textbf{The layer to tune ${\textsc{L}}$}.
We now investigate the impact of layer selection on the performance of our FL and FLLoRA attacks. All target models in our study follow the transformer encoder-decoder architecture \citep{vaswani2017attention}, where each component consists of a stack of attention layers, and a shared embedding projection layer maps the hidden states to logit vectors for prediction. Given this common structure, we examine the effect of tuning similar layers across all models, with results for attack accuracy presented in Table \ref{tab:ablate_layer}.

\begin{table}[t]
\begin{center}
\begin{small}
\small
\begin{tabular}{lccc}
\toprule
Layer & VT5(PFL) & Donut(DocVQA) & Pix2Struct-B(DocVQA)\\
\midrule 
Embedding Projection Layer & 67.0 & 71.33 & 68.66\\
Embedding Layer Norm & 65.33 & 76.0 & 64.67\\
\cmidrule{1-4}
Last Decoder Block FC1 & \textbf{68.33} & \textbf{78.0} & 68\\
Last Decoder Block FC2 & 68.17 & 77.33 & \textbf{68.83}\\
Last Decoder Block Layer Norm & 61.83 & 76.83 & 67.5\\
\cmidrule{1-4}
Random Decoder Block FC1 & 61.33 & 72.0 & 67.5\\
Random Decoder Block FC2 & 64.0 & 73.0 & 65.17\\
\bottomrule
\end{tabular}
\end{small}
\end{center}
\caption{\textbf{Effect of selected layer to tune} from each target model. Attack performances are reported in terms of Accuracy.}
\label{tab:ablate_layer}
\vspace{-0.2in}
\end{table}

Our findings reveal that \textit{layers closer to the final output exhibit higher privacy leakage} in terms of MI compared to (randomly selected) intermediate layers, likely due to receiving larger gradient updates. Specifically, fine-tuning the final fully connected layer alone leads to strong attack performance while also being more efficient in terms of the number of parameters that need to be optimized. This suggests that focusing on the last layers can achieve both high privacy leakage and computational efficiency in our MI attacks.

\textbf{Threshold $\tau$}.
With the optimizer $\textsc{OPT}$ and learning rate $\alpha$ fixed, the threshold $\tau$ emerges as the most critical hyperparameter that requires careful tuning for each attack. We experiment with a wide range of $\tau$ values, spanning from $10^{-6}$ to $10.0$, and select the optimal value based on attack performance, as demonstrated in Figure \ref{fig:ablate_tau}. This optimal $\tau$ is then applied consistently in all subsequent experiments. Careful selection of this threshold is crucial, as it directly influences the stability and success of the optimization process.

\begin{table}[t]
\begin{center}
\begin{small}
\small
\begin{tabular}{ccccccc}
\toprule
Model & $\alpha_{\textsc{FL}}$ & $\alpha_{\textsc{IG}}$ & $S$ & $L$ & $\tau_{\text{FL}}$ & $\tau_{\text{IG}}$\\
\midrule 
VT5 & \multirow{3}{*}{0.001} & 1.0 & \multirow{3}{*}{200} & \multirow{3}{*}{last FC layer} & $10^{-6}$ & $10^{-5}$\\
Donut &  & 0.001 &  &  & 1.0 & 5.0\\
Pix2Struct-B &  & 0.001 &  &  & $10^{-4}$ & $10^{-3}$\\
\bottomrule
\end{tabular}
\end{small}
\end{center}
\caption{\textbf{Best Hyperaremeters from our tuning process} with consistent performance across both PFL and DocVQA dataset.}
\label{tab:best_hps}
\end{table}

\begin{table}[t]
\begin{center}
\begin{small}
\begin{adjustbox}{width=1\textwidth}
\small
\begin{tabular}{ccccccc}
\toprule
\multirow{2}{*}{Model} & \multirow{2}{*}{Num. Params} & \multirow{2}{*}{Downstream Task} & \multicolumn{2}{c}{Data} & \multicolumn{2}{c}{Checkpoint} \\
\cmidrule(l){4-5}
\cmidrule(l){6-7}
& & & Pretrain & Finetune & Pretrain & Finetune \\
\midrule 
\multirow{2}{*}{VT5} & \multirow{2}{*}{250M} & \multirow{2}{*}{DocVQA} & \multirow{2}{*}{C4+IIT-CDIP} & PFL & \multicolumn{2}{c}{\multirow{2}{*}{https://benchmarks.elsa-ai.eu/?ch=2}} \\
 &  &  &  & DocVQA &  & \\
\midrule 

\multirow{2}{*}{Donut} & \multirow{2}{*}{200M} & \multirow{2}{*}{DocVQA} & \multirow{2}{*}{CDIP 11M + 0.5M synthesized Docs} & PFL &  \multicolumn{2}{c}{Ours}  \\
 &  &  &  & DocVQA & $\text{naver-clova-ix/donut-base}^{\dagger}$ & $\text{naver-clova-ix/donut-base-finetuned-docvqa}^{\dagger}$  \\
\midrule 
Pix2struct-B & 282M & \multirow{2}{*}{DocVQA} & \multirow{2}{*}{BooksCorpus + C4 Web HTML} & \multirow{2}{*}{DocVQA} & $\text{google/pix2struct-base}^{\dagger}$ & $\text{google/pix2struct-docvqa-base}^{\dagger}$ \\
Pix2struct-L & 1.33B &  &  &  & $\text{google/pix2struct-large}^{\dagger}$ & $\text{google/pix2struct-docvqa-large}^{\dagger}$ \\
\bottomrule
\end{tabular}
\end{adjustbox}
\end{small}
\end{center}
\caption{\textbf{Details of the public checkpoints} used as target models in this work. $\dagger$ denotes checkpoint from Hugging Face.}
\label{tab:public_checkpoint}
\vskip -0.1in
\end{table}

We summarize the set of tuned hyperparameters for our approach in Table \ref{tab:best_hps}.

\section{More on Attack Implementation}
\label{sec:attack_implementation}
\subsection{Target Model Training}

For all target models, whenever feasible, we utilize the public checkpoint fine-tuned on the considered private dataset from Hugging Face library and adhere to the data processing guidelines, such as document resolution, as recommended by the authors. We deliberately opt for public checkpoints for two reasons: (1) to make it consistent to further research in privacy attacks that use the same trained models, and (2) to minimizing the biases in model training that affect the final results, given the complexity of the original training process and our limited resources. Table \ref{tab:public_checkpoint} summarizes the details of the process from which public checkpoints for the target models considered in this work are obtained. This includes the datasets the models were pre-trained on, before by fine-tuning on target DocVQA datasets, along with the corresponding download URLs for these checkpoints.

\begin{table}[h]
\begin{center}
\begin{small}
\begin{adjustbox}{width=1\textwidth}
\small
\begin{tabular}{ccccccc}
\toprule
Model & Optimizer & Learning Rate & Weight Decay & Batch Size & Scheduler & Iteration \\
\midrule
VT5 & AdamW & 2e-4 & - & 16 & - & 200k\\
Donut & Adam & 3e-5 & 0.01 & 4 & linear warmup 10\% & 800k\\
Pix2Struct-B & AdaFactor & 1e-5 & - & 4 & warmup 1000 steps, cosine decay to 0 & 800k\\
\bottomrule
\end{tabular}
\end{adjustbox}
\end{small}
\end{center}
\caption{\textbf{Details of the training hyperparameter} for each target model in this work.}
\label{tab:training_params}
\vskip -0.1in
\end{table}

If public checkpoints are unavailable, we fine-tune the selected model on the respective private dataset, using the pre-trained checkpoint as the initialization, with the training procedure outlined by the respective authors. To prevent overfitting, we perform early stopping based on validation performance, ensuring that all evaluated models generalize well to unseen data. We also use the pre-trained checkpoint to initialize the proxy model $\mathcal{F}_p$ to train it on $D_{\text{query}}$. We provide an overview of the training procedure for each target model, based on the respective papers. These procedures were adapted to fit our computational resources, as outlined in Table \ref{tab:training_params}.
\begin{table}[t]
\begin{center}
\begin{small}
\small
\begin{adjustbox}{width=0.6\textwidth}
\begin{tabular}{lccccc}
\toprule
Dataset & Model & Test Set & ACC & ANLS & Train-Test Gap \\
\midrule
\multirow{6}{*}{{PFL}} & \multirow{3}{*}{{VT5}} & Original &  81.4 & 90.17  & -\\
 &  & MIA & 82.74 & 90.91 & 11.44 \\
 &  & MIA-rephrased & 77.59 & 85.84 & -\\
\cmidrule{2-6}
& \multirow{3}{*}{{Donut}} & Original & 74.73 & 88.66 & -\\
 &  & MIA & 80.15 & 91.64 & 22.2 \\
 &  & MIA-rephrased & 70.46 & 80.96 & -\\
\midrule
\multirow{12}{*}{{DVQA}}& \multirow{3}{*}{{VT5}} & Original & 60.1 & 69.33 & -\\
 &  & MIA & 75.54 & 81.69 & 36.22 \\
 &  & MIA-rephrased & 73.57 & 79.89 & -\\
\cmidrule{2-6}
& \multirow{3}{*}{{Donut}} & Original & 59.26 & 66.91 & -\\
 &  & MIA & 78.55 & 83.42 & 39.78 \\
 &  & MIA-rephrased & 72.57 & 77.12 & -\\
\cmidrule{2-6}
& \multirow{3}{*}{{Pix2Struct-B}} & Original & 57.11 & 68.13 & -\\
 &  & MIA & 64.42 & 79.95 & 25.8 \\
 &  & MIA-rephrased & 63.81 & 74.06 & -\\
\cmidrule{2-6}
& \multirow{3}{*}{{Pix2Struct-L}} & Original & 64.53 & 74.12 & -\\
 &  & MIA & 73.91 & 82.71 & 22.11 \\
 &  & MIA-rephrased & 69.93 & 79.15 & -\\
\bottomrule
\end{tabular}
\end{adjustbox}
\end{small}
\end{center}
\caption{\textbf{DocVQA Performance of the target models on PFL and DocVQA dataset.} Train-Test Gap is computed as the different of DocVQA Accuracy between \textit{member/non-member} documents. $\textsc{MIA}$ denotes the attack evaluation set, which is a subset randomly sampled from the original train/test set, $\textsc{MIA}\text{-rephrased}$ is its variants with rephrased questions by LLM.}
\label{tab:docvqa_performance}
\end{table}

\subsection{Target Model Performance on DocVQA}
To ensure the utility of the target models for our experiments, we validated that the DocVQA performance of each model checkpoint closely matched the results reported in the respective papers. Table \ref{tab:docvqa_performance}  presents the target models' performance across both DocVQA datasets. We observe a clear train-test performance gap, particularly in smaller models, while the gap tends to narrow for more generalized models or with increased dataset size.

\subsection{Computation and Runtime} All attack methods are implemented using PyTorch and executed on an NVIDIA GeForce A40 GPU with 45 GB of memory. The maximum runtime for each attack does not exceed \textit{10 hours} per run, depending on the target model’s size and the preprocessing steps required for the data. This runtime reflects the efficiency of our approach, especially when compared to methods based on shadow training, which require retraining of large-scale models many times to be effective \citep{carlini2022membership}. Our results demonstrate that the proposed attacks are both efficient and scalable, making them practical for large-scale models in real-world applications.

\section{More on Attack Results}
\begin{table}[h]
\begin{center}
\begin{small}
\begin{adjustbox}{width=1\textwidth}
\small
\begin{tabular}{clcccccc}
\toprule
\multicolumn{2}{c}{\multirow{2}{*}{{White-box}}} & \multicolumn{2}{c}{FL} & \multicolumn{2}{c}{FLLoRA} & \multicolumn{2}{c}{IG}\\
\cmidrule(l){3-4}
\cmidrule(l){5-6}
\cmidrule(l){7-8}
&                                 & ACC & F1 & ACC & F1 & ACC & F1 \\
\midrule 
\multirow{2}{*}{\rotatebox[origin=c]{90}{PFL}}& {VT5} & $\textbf{66.63}_{\textbf{0.07}}\textbf{\textcolor{red}{(+5.96)}}$ & $\textbf{72.40}_{\textbf{0.1}}\textbf{\textcolor{red}{(+11.73)}}$ & $65.17_{0.0}\textcolor{red}{(+4.50)}$ & $70.52_{0.0}\textcolor{red}{(+9.85)}$ & $61.83_{0.0}\textcolor{red}{(+1.16)}$ & $69.18_{0.0}\textcolor{red}{(+8.51)}$\\

& {Donut} & $72.67_{0.0}\textcolor{red}{(+1.5)}$ & $77.96_{0.0}\textcolor{red}{(+6.63)}$ & $72.83_{0.0}\textcolor{red}{(+1.66)}$ & $77.94_{0.0}\textcolor{red}{(+6.61)}$ & $\textbf{75.17}_{\textbf{0.0}}\textbf{\textcolor{red}{(+4)}}$ & $\textbf{79.39}_{\textbf{0.0}}\textbf{\textcolor{red}{(+8.06)}}$\\
\midrule

\multirow{3}{*}{\rotatebox[origin=c]{90}{DVQA}}& {VT5} & $75.67_{0.0}\textcolor{red}{(+0.5)}$ & $76.60_{0.0}\textcolor{red}{(+1.47)}$ & $75.57_{0.08}\textcolor{red}{(+0.4)}$ & $77.31_{0.13}\textcolor{red}{(+2.18)}$ & $74.83_{0.0}\textcolor[HTML]{0b5394}{(-0.34)}$ & $76.88_{0.0}\textcolor{red}{(+1.75)}$\\

& {Donut} & $80.33_{0.0}\textcolor[HTML]{0b5394}{(-0.17)}$ & $82.18_{0.0}\textcolor{red}{(+1.08)}$ & $80.0_{0.0}\textcolor[HTML]{0b5394}{(-0.5)}$ & $81.93_{0.0}\textcolor{red}{(+0.83)}$ & $80.33_{0.0}\textcolor[HTML]{0b5394}{(-0.17)}$ & $82.34_{0.0}\textcolor{red}{(+1.24)}$\\

& {Pix2Struct-B} & $71.67_{0.0}\textcolor{red}{(+2.54)}$ & $72.22_{0.0}\textcolor{red}{(+4.55)}$ & $70.50_{0.0}\textcolor{red}{(+1.37)}$ & $71.95_{0.0}\textcolor{red}{(+4.28)}$ & $\textbf{72.00}_{\textbf{0.0}}\textbf{\textcolor{red}{(+2.87)}}$ & $\textbf{72.82}_{\textbf{0.0}}\textbf{\textcolor{red}{(+5.15)}}$\\
\bottomrule
\end{tabular}
\end{adjustbox}
\end{small}
\end{center}
\vskip -0.1in
\caption{\textbf{White-Box: Main Results of \atk.} Values in parentheses indicate the improvement (\textcolor{red}{positive}/\textcolor[HTML]{0b5394}{negative}) of our proposed attacks compared to the $\textsc{ScoreLoss-UA}_{\text{all}}$. Compared to all baselines, the methods with the best \textit{average} performance across the two metrics are highlighted in \textbf{bold}. Results are reported over five random seeds.}
\label{tab:whitebox_results}
\end{table}

\label{sec:tpr_at_fix_fpr}
In this section, we evaluate our attacks using TPR@1\%FPR and TPR@3\%FPR, following standard practices in recent MIA literature. The results are summarized in Table~\ref{tab:tpr@fpr_whitebox} and ~\ref{tab:tpr@fpr_blackbox}. 

\begin{table}[t]
\begin{center}
\begin{small}
\begin{adjustbox}{width=0.96\textwidth}
\small
\begin{tabular}{lcccccccccc}
\toprule
 & \multicolumn{6}{c}{\textbf{DVQA}} & \multicolumn{4}{c}{\textbf{PFL}}  \\
\cmidrule(l){2-7}
\cmidrule(l){8-11}
 & \multicolumn{2}{c}{VT5} & \multicolumn{2}{c}{Donut} & \multicolumn{2}{c}{Pix2Struct-B} & \multicolumn{2}{c}{VT5} & \multicolumn{2}{c}{Donut} \\
\cmidrule(l){2-3}
\cmidrule(l){4-5}
\cmidrule(l){6-7}
\cmidrule(l){8-9}
\cmidrule(l){10-11}
 & 1\% & 3\% & 1\% & 3\% & 1\% & 3\% & 1\% & 3\% & 1\% & 3\% \\
\midrule
$\textsc{Loss-TA}$ & \textbf{7.67} & \textbf{14.00} & 0.67 & 7.67 & 2.33 & 5.33 & 0.67 & 3.00 & 1.67 & \textbf{14.67} \\
$\textsc{Gradient-UA}$ & 2.33 & 9.33 & 3.67 & 6.00 & 1.00 & 5.00 & 0.33 & 3.00 & 1.00 & 8.33 \\
$\textsc{ScoreLoss-UA}_{\text{all}}$ & 1.33 & 4.67 & 2.67 & 8.67 & 2.00 & 6.67 & 0.33 & 4.00 & 0.67 & 6.33 \\
Min-K\% & 2.67 & 10.67 & 0.33 & 1.33 & 0.33 & 5.33 & 1.67 & 5.67 & 0.00 & 0.00 \\
Min-K\%++ & 1.00 & 7.00 & \textbf{4.33} & 9.33 & 0.67 & 10.33 & 1.00 & 8.00 & 0.33 & 2.00 \\
\midrule
FL & 2.33 & 5.67 & 3.33 & \textbf{10.67} & \textbf{6.00} & \textbf{11.00} & \textbf{3.67} & \textbf{8.67} & 0.33 & 7.00 \\
FLLoRA & 3.33 & 11.33 & 2.67 & 5.33 & 3.67 & 6.33 & 1.33 & 3.33 & 0.33 & 10.00 \\
IG & 0.67 & 5.67 & 1.33 & 8.00 & 3.00 & 10.33 & 1.00 & 2.33 & \textbf{5.67} & 11.00 \\
\bottomrule
\end{tabular}
\end{adjustbox}
\end{small}
\end{center}
\vskip -0.1in
\caption{\textbf{White-box: TPR at fixed FPR}. Comparison across all white-box methods, with the best-performance highlighted in \textbf{bold}. 1\% and 3\% indicate TPR@1\%FPR and TPR@3\%FPR respectively.}
\label{tab:tpr@fpr_whitebox}
\end{table}

\begin{figure}[t]
    \centering
    \begin{minipage}[b]{0.66\textwidth}
        \centering
        \resizebox{0.999\linewidth}{!}{
            \includegraphics[height=5cm,width=1.05\textwidth]{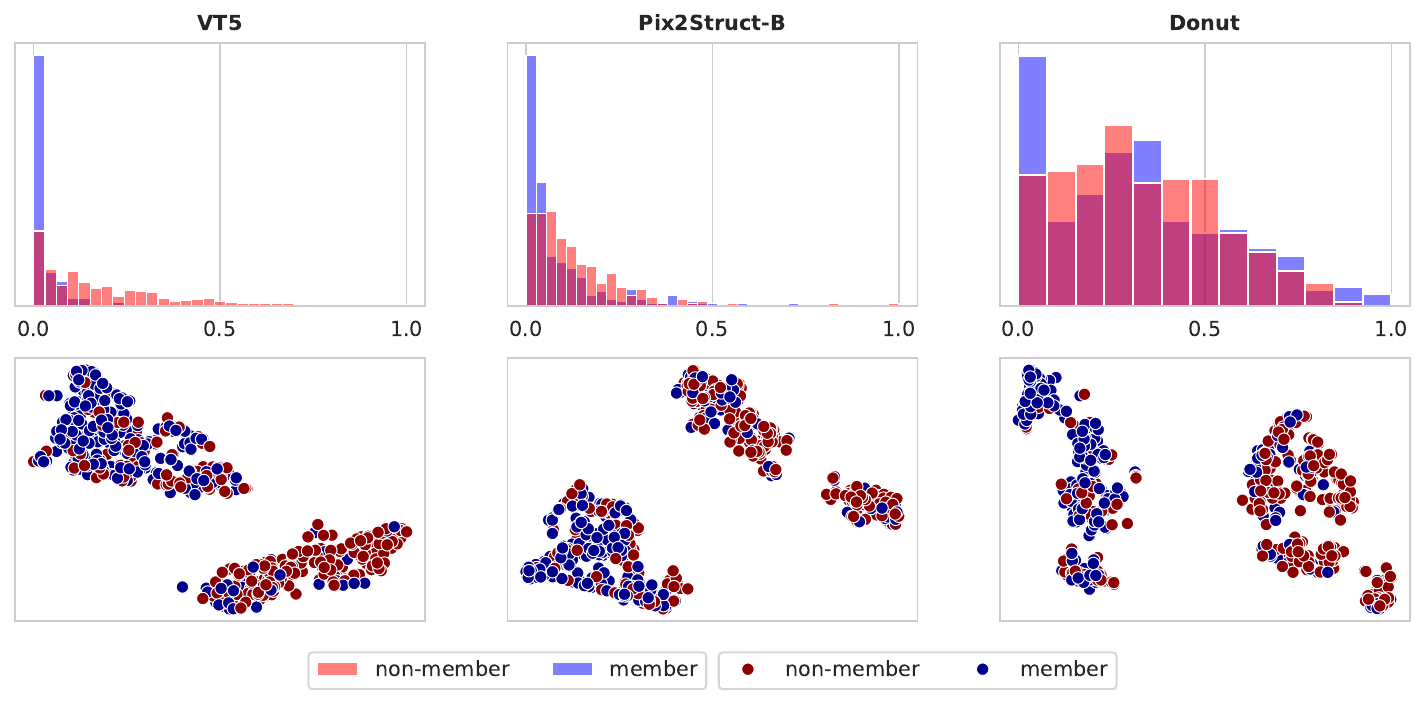}
        }
        \caption{\textbf{Membership Features against three different target models on DocVQA Dataset.} \textit{Top}: The distribution of average \textbf{\textit{loss}} over all questions from all target documents on each target model. \textit{Bottom}: T-SNE visualization of the features used in our proposed attacks.}
        \label{fig:whitebox_loss}
    \end{minipage}
    \hfill
    \begin{minipage}[b]{0.3\textwidth}
        \centering
        \includegraphics[height=5cm]{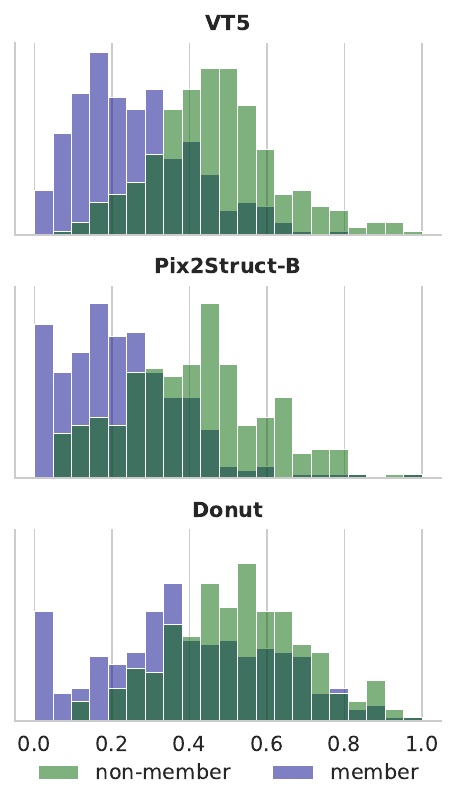}
        \caption{\textbf{Distribution of Average \textit{Distance}} from member and non-member documents.}
        \label{fig:whitebox_distance}
    \end{minipage}
\label{fig:whitebox_feature}
\vspace{-0.2in}
\end{figure}

An interesting observation is the high performance of the \textsc{LOSS-TA} method for VT5 on DocVQA and Donut on PFL in Table~\ref{tab:tpr@fpr_whitebox}. This performance can be attributed to the clear separation in the loss distribution between member and non-member samples (Figure~\ref{fig:whitebox_loss}), which indicates overfitting behavior in these cases.

\section{More on Analysis}
\label{sec:more_analysis}
In this section, we provide a deeper analysis of the effectiveness of our proposed white-box and black-box attacks, highlighting their performance relative to the baseline approaches.

\subsection{Impact of Selected Features}
\label{sec:impact_feature}
As outlined in the main paper, we fix the set of selected features across all experiments. These features include the DocVQA score $u$, the optimization-based distance $\Delta$, and the number of optimization steps $s$, aggregated using the set of aggregation functions $\Phi_\text{all}=\{\textsc{avg}; \textsc{min}; \textsc{max}; \textsc{med}\}$ . We first evaluate the impact of individual features and their combinations on attack performance in the white-box DocMIA setting, using $\textsc{avg}$ as the aggregation function $\Phi$. The analysis employs the best hyperparameters identified during the tuning process described in Section \ref{sec:calibration}.

Table \ref{tab:impact_feature_pfl} and Table \ref{tab:impact_feature_docvqa} summarize the attack performance when individual features or their combinations are used. 
Additionally, Table \ref{tab:impact_feature_more} \textit{(Top)} compares the attack performance of our optimization-based features with the loss value $\ell$ and the gradient norm of the loss with respect to the model parameters $\theta$. Here, the loss value $\ell$ is computed uniformly across all target models over $K$ generation steps, given a (document, question, answer) example $(x,q,a)$ as:
\begin{equation}
    \ell =-\sum^{K}_{k=1} \log{p_{\theta} (a_k|a_{<k},x,q)}
\end{equation}
When used individually, our proposed optimization-based features outperform the DocVQA score and the loss in most cases. Our attack methods are particularly effective against target models like VT5 and Donut trained on PFL-DocVQA, which exhibit lower overfitting and small Train-Test gaps (as shown in Table\ref{tab:docvqa_performance}). These results highlight that our attacks provide more discriminative features than the commonly used MIA features.

When combined, our selected features achieve the best or near-best performance across all cases. Furthermore, extending aggregation functions from $\textsc{avg}$ to $\Phi_\text{all}$ adds notable improvements in attack effectiveness, as shown in Table \ref{tab:impact_feature_more} \textit{(Bottom)}. These results demonstrate that our proposed feature set is robust across different target models, making it a reliable choice for DocMIA.
\begin{table}[t]
    \centering
    \begin{minipage}{0.64\textwidth}
        \centering
        \resizebox{\linewidth}{!}{
            \begin{tabular}{cc}
                \begin{tabular}{cccc}
                    \toprule
                    \multicolumn{4}{c}{VT5} \\
                    \midrule
                    $\textsc{avg}(\textsc{nls})$ & $\textsc{avg}(\Delta)$ & $\textsc{avg}(s)$ & F1\\
                    \midrule 
                    \checkmark &  &  & 68.88\\
                     & \checkmark &  & 71.45\\
                     &  & \checkmark & 70.92\\
                     \midrule
                    \checkmark & \checkmark &  & 71.09\\
                    \checkmark &  & \checkmark & 71.11\\
                     & \checkmark & \checkmark & 71.22\\
                    \midrule
                    \checkmark & \checkmark & \checkmark & {\textbf{71.53}}\\
                    \bottomrule
                    \label{tab:impact_feature_pfl_vt5}
                \end{tabular}
                &
                \begin{tabular}{cccc}
                    \toprule
                    \multicolumn{4}{c}{Donut} \\
                    \midrule
                    $\textsc{avg}(\textsc{nls})$ & $\textsc{avg}(\Delta)$ & $\textsc{avg}(s)$ & F1\\
                    \midrule
                    \checkmark &  &  & 67.58\\
                     & \checkmark &  & 71.36\\
                     &  & \checkmark & 73.16\\
                    \midrule
                    \checkmark & \checkmark &  & 72.87\\
                    \checkmark &  & \checkmark & 73.67\\
                     & \checkmark & \checkmark & 73.86\\
                    \midrule
                    \checkmark & \checkmark & \checkmark & {\textbf{73.89}}\\
                    \bottomrule
                \label{tab:impact_feature_pfl_donut}
                \end{tabular}
            \end{tabular}
        }
        \caption{\textbf{Impact of Selected Features on PFL-DocVQA Models.}}
        \label{tab:impact_feature_pfl}
    \end{minipage}
    \hfill
    \begin{minipage}{0.35\textwidth}
        \centering
        \resizebox{\linewidth}{!}{
            \begin{tabular}{c}
                \begin{tabular}{cccccc}
                    \toprule
                     & \multicolumn{2}{c}{PFL} & \multicolumn{3}{c}{DVQA}\\
                    \cmidrule(l){2-3}
                    \cmidrule(l){4-6}
                     & VT5 & Donut & VT5 & Donut & Pix2Struct-B\\
                    \midrule 
                    $\textsc{avg}(\ell)$ & 67.53 & 67.80 & 73.43 & 56.79 & 69.97\\
                    $\textsc{avg}(||\nabla_{\theta}\mathcal{L}||_2)$ & 70.53 & 71.51 & 71.91 & 71.53 & 66.14\\
                    $\textsc{avg}(\Delta)$ & 71.45 & 71.36 & 72.86 & 57.34 & 70.57\\
                    $\textsc{avg}(s)$ & 70.92 & 73.16 & 74.34 & 60.32 & 69.00\\
                    \bottomrule
                    \\
                    \toprule
                     & \multicolumn{2}{c}{PFL} & \multicolumn{3}{c}{DVQA}\\
                    \cmidrule(l){2-3}
                    \cmidrule(l){4-6}
                     & VT5 & Donut & VT5 & Donut & Pix2Struct-B\\
                    \midrule
                    $\Phi=\textsc{avg}$ & 71.53 & 73.89 & 74.96 & 72.94 & 73.22\\
                    $\Phi=\Phi_{\text{all}}$ & 72.4\textcolor{red}{(+0.87)} & 77.96\textcolor{red}{(+4.07)} & 76.6\textcolor{red}{(+1.67)} & 82.18\textcolor{red}{(+9.24)} & 72.22\textcolor{blue}{(-1.0)}\\
                    \bottomrule							
                \end{tabular}
            \end{tabular}
        }
    \caption{\textbf{Comparisons in Attack Performance in terms of F1 Score}: \textit{(Top)} between our Optimization-based Features with the loss value $\ell$ and the gradient norm $||\nabla_{\theta}\mathcal{L}||_2$. \textit{(Bottom)} between $\textsc{avg}$ and $\Phi_{\text{all}}$ as the aggregation functions.}
    \label{tab:impact_feature_more}
    \end{minipage}
\end{table}

\begin{table}[t]
    \centering
    \begin{minipage}{0.32\textwidth}
        \centering
        \resizebox{\linewidth}{!}{%
            \begin{tabular}{cccc}
                \toprule
                \multicolumn{4}{c}{VT5} \\
                \midrule
                $\textsc{avg}(\textsc{nls})$ & $\textsc{avg}(\Delta)$ & $\textsc{avg}(s)$ & F1\\
                \midrule 
                \checkmark &  &  & 72.73\\
                 & \checkmark &  & 72.86\\
                 &  & \checkmark & 74.34\\
                 \midrule
                \checkmark & \checkmark &  & \textbf{75.81}\\
                \checkmark &  & \checkmark & 75.04\\
                 & \checkmark & \checkmark & 74.19\\
                \midrule
                \checkmark & \checkmark & \checkmark & 74.96\\
                \bottomrule
            \end{tabular}
        }
        \label{tab:impact_feature_docvqa_vt5}
    \end{minipage}
    \begin{minipage}{0.32\textwidth}
        \centering
        \resizebox{\linewidth}{!}{%
            \begin{tabular}{cccc}
                \toprule
                \multicolumn{4}{c}{Donut} \\
                \midrule
                $\textsc{avg}(\textsc{nls})$ & $\textsc{avg}(\Delta)$ & $\textsc{avg}(s)$ & F1\\
                \midrule 
                \checkmark &  &  & \textbf{76.88}\\
                 & \checkmark &  & 57.34\\
                 &  & \checkmark & 60.32\\
                \midrule
                \checkmark & \checkmark &  & 65.94\\
                \checkmark &  & \checkmark & 72.17\\
                 & \checkmark & \checkmark & 60.29\\
                \midrule
                \checkmark & \checkmark & \checkmark & 72.94\\
                \bottomrule
            \end{tabular}
        }
        \label{tab:impact_feature_docvqa_donut}
    \end{minipage}
    \begin{minipage}{0.32\textwidth}
        \centering
        \resizebox{\linewidth}{!}{%
            \begin{tabular}{cccccc}
                \toprule
                \multicolumn{4}{c}{Pix2Struct-B} \\
                \midrule
                $\textsc{avg}(\textsc{nls})$ & $\textsc{avg}(\Delta)$ & $\textsc{avg}(s)$ & F1\\
                \midrule 
                \checkmark &  &  & 72.60\\
                 & \checkmark &  & 70.57\\
                 &  & \checkmark & 69.00\\
                \midrule
                \checkmark & \checkmark &  & 73.20\\
                \checkmark &  & \checkmark & 72.87\\
                 & \checkmark & \checkmark & 70.17\\
                \midrule
                \checkmark & \checkmark & \checkmark & \textbf{73.22}\\
                \bottomrule
            \end{tabular}
        }
        \label{tab:impact_feature_docvqa_pix2struct}
    \end{minipage}
\caption{\textbf{Impact of Selected Features on DocVQA Target Models}. Only $\textsc{AVG}$ is used as the aggregation function $\Phi$. Attack performances are obtained with our \textsc{FL} method using the best hyperparameters.}
\label{tab:impact_feature_docvqa}
\end{table}
\subsection{Impact of the Training Questions Knowledge}
\label{sec:impact_question_knowledge}
So far, our document MI attacks against DocVQA models have assumed complete knowledge of the original training questions. We now relax this assumption and investigate how the lack of access to the exact training questions affects attack performance. In practice, an adversary may not have access to the exact training questions but can approximate them. For example, documents like invoices often follow standard layouts, and biases in human annotation may lead to predictable patterns in the types of questions asked during the creation of DocVQA datasets \citep{tito2024privacy,mathew2021docvqa}. It is important to note that the original training questions tend to be simple, natural questions designed to extract specific information from the document. Moreover, the type of question is inherently linked to the type of document on which the DocVQA model is trained. For instance, if the target model is trained on invoices, the natural type of question would focus on extracting essential details from the invoice, such as the “total amount”, framed in a clear and straightforward manner e.g., "What is the total?".
This makes it possible for an adversary to generate approximate versions of the training questions, simulating a more realistic attack setting.
\begin{table}[t]
\begin{center}
\begin{small}
\begin{adjustbox}{width=1\textwidth}
\small
\begin{tabular}{clcccccccccccc}
\toprule
 & \multicolumn{1}{c}{\multirow{2}{*}{\textbf{Target}}} & \multicolumn{8}{c}{\textbf{DVQA}} & \multicolumn{4}{c}{\textbf{PFL}} \\
\cmidrule(l){3-10}
\cmidrule(l){11-14}
 &  & \multicolumn{2}{c}{\textbf{VT5}} & \multicolumn{2}{c}{\textbf{Donut}} & \multicolumn{2}{c}{\textbf{Pix2Struct-B}} & \multicolumn{2}{c}{\textbf{Pix2Struct-L}} & \multicolumn{2}{c}{\textbf{VT5}} & \multicolumn{2}{c}{\textbf{Donut}} \\
\cmidrule(l){3-4}
\cmidrule(l){5-6}
\cmidrule(l){7-8}
\cmidrule(l){9-10}
\cmidrule(l){11-12}
\cmidrule(l){13-14}
Proxy &  & 1\% & 3\% & 1\% & 3\% & 1\% & 3\% & 1\% & 3\% & 1\% & 3\% & 1\% & 3\%  \\
\midrule
& {$\textsc{Score-TA}$} & 4.00 & 9.33 & 5.00 & 11.00 & \textbf{5.33} & 8.00 & 3.33 & \textbf{9.00} & 1.00 & 5.00 & 0.67 & 2.67 \\
& {$\textsc{Score-UA}$} & 3.67 & 7.67 & 4.33 & 15.67 & 4.00 & 6.33 & 4.33 & 6.67 & 0.67 & 3.33 & 0.33 & 3.33 \\
& {$\textsc{Score-UA}_{\text{all}}$} & 4.00 & 9.33 & 5.00 & 11.00 & 5.33 & 8.00 & 3.33 & 9.00 & 1.00 & 5.00 & 0.67 & 2.67 \\
\midrule
\multirow{3}{*}{VT5} & FL & 0.67 & \textbf{12.33} & \textbf{11.67} & \textbf{23.00} & 2.00 & \textbf{16.67} & 2.00 & 5.33 & 0.67 & 2.00 & \textbf{5.00} & \textbf{8.00} \\
& FLLoRA & \textbf{4.67} & {11.33} & 6.34 & 16.33 & 2.33 & 9.33 & 1.00 & 4.67 & 2.00 & 3.33 & 0.00 & 2.00 \\
& IG & 1.00 & 8.33 & 2.00 & 7.00 & 4.67 & 7.67 & 2.33 & 7.00 & 0.33 & 3.67 & 1.33 & 6.67 \\
\cmidrule(l){1-14}
\multirow{3}{*}{Donut} & FL & 0.33 & 6.33 & 0.33 & 4.00 & 1.33 & 4.67 & 3.00 & 7.33 & 0.33 & 1.33 & 1.33 & 4.00 \\
& FLLoRA & 1.00 & 6.33 & 1.67 & 5.00 & 2.33 & 6.33 & 3.00 & 8.00 & 0.00 & 5.33 & 2.00 & 5.33 \\
& IG & 1.67 & 5.00 & 0.67 & 11.00 & 3.67 & 9.33 & \textbf{4.67} & 6.33 & \textbf{2.67} & \textbf{6.33} & 1.67 & 4.33 \\
\bottomrule
\end{tabular}
\end{adjustbox}
\end{small}
\end{center}
\vskip -0.1in
\caption{\textbf{Black-box: TPR at fixed FPR}. Comparison across all black-box methods, with the best-performing method highlighted in \textbf{bold}. 1\% and 3\% indicate TPR@1\%FPR and TPR@3\%FPR respectively.}
\label{tab:tpr@fpr_blackbox}
\end{table}

\begin{table}[h]
\vskip 0.15in
\begin{center}
\begin{small}
\begin{adjustbox}{width=1\textwidth}
\small
\begin{tabular}{clcccccccccc}
\toprule
\multicolumn{2}{c}{\multirow{2}{*}{Model}} & \multicolumn{2}{c}{$\textsc{Score-TA}$} & \multicolumn{2}{c}{$\textsc{Score-UA}_{\text{all}}$} & \multicolumn{2}{c}{$\textsc{Loss-TA}$} & \multicolumn{2}{c}{$\textsc{ScoreLoss-UA}_{\text{all}}$} & \multicolumn{2}{c}{$\textsc{Ours (FL)}$}\\
\cmidrule(l){3-4}
\cmidrule(l){5-6}
\cmidrule(l){7-8}
\cmidrule(l){9-10}
\cmidrule(l){11-12}
&                                 & ACC & F1 & ACC & F1 & ACC & F1 & ACC & F1 & ACC & F1 \\
\midrule 
\multirow{2}{*}{\rotatebox[origin=c]{90}{\tiny PFL}} & VT5 & \cellcolor[HTML]{C0C0C0}{$60.67$} & \cellcolor[HTML]{C0C0C0}{$64.13$} & \cellcolor[HTML]{C0C0C0}{$55.83_{0.0}$} & \cellcolor[HTML]{C0C0C0}{$46.89_{0.0}$} & $54.50$ & $59.19$ & $55.83_{0.0}$ & $46.89_{0.0}$ & $\textbf{\textcolor{red}{64.00}}_{\textbf{\textcolor{red}{0.0}}}$ & ${\textbf{\textcolor{red}{69.14}}}_{\textbf{\textcolor{red}{0.0}}}$\\

& {Donut} & \cellcolor[HTML]{C0C0C0}{$69.17$} & \cellcolor[HTML]{C0C0C0}{$69.72$} & \cellcolor[HTML]{C0C0C0}{$59.33_{0.0}$} & \cellcolor[HTML]{C0C0C0}{$51.59_{0.0}$} & $68.50$ & $66.67$ & $59.17_{0.0}$ & $51.49_{0.0}$ & $\textbf{\textcolor{red}{71.13}}_{\textbf{\textcolor{red}{0.08}}}$ & $\textbf{\textcolor{red}{72.07}}_{\textbf{\textcolor{red}{0.0}}}$\\
\midrule

\multirow{2}{*}{\rotatebox[origin=c]{90}{\tiny DVQA}} & VT5 & \cellcolor[HTML]{C0C0C0}{$73.67$} & \cellcolor[HTML]{C0C0C0}{$75.01$} & \cellcolor[HTML]{C0C0C0}{$74.83_{0.0}$} & \cellcolor[HTML]{C0C0C0}{$74.36_{0.0}$} & $71.67$ & $74.06$ & $75.17_{0.0}$ & $74.96_{0.0}$ & $\textbf{\textcolor{red}{74.83}}_{\textbf{\textcolor{red}{0.0}}}$ & $\textbf{\textcolor{red}{75.68}}_{\textbf{\textcolor{red}{0.0}}}$\\
& {Donut} & \cellcolor[HTML]{C0C0C0}{\textbf{\textcolor{red}{69.17}}} & \cellcolor[HTML]{C0C0C0}{$\textbf{\textcolor{red}{71.23}}$} & \cellcolor[HTML]{C0C0C0}{$65.17_{0.0}$} & \cellcolor[HTML]{C0C0C0}{$62.21_{0.0}$} & $52.33$ & $53.57$ & $65.17_{0.0}$ & $62.21_{0.0}$ & $67.67_{0.0}$ & $68.51_{0.0}$\\
\bottomrule
\end{tabular}
\end{adjustbox}
\end{small}
\end{center}
\vskip -0.1in
\caption{\textbf{Results with Rephrased Questions.} \textcolor{gray}{\textbf{Gray}} color indicate attacks conducted in the black-box setting. All results are reported based on five random seeds. The methods with the best \textit{average} performance across the two metrics are highlighted in \textbf{\textcolor{red}{bold}}.}
\label{tab:rephrased_question_results}
\end{table}

To explore this scenario, we conduct experiments where we paraphrase the original training questions using Mistral~\citep{jiang2023mistral}, and use these rephrased questions as inputs for the MI attacks. As illustrated in Table \ref{tab:rephrased_question_results}, the performance of all MI attacks declines when rephrased questions are used, mirroring the drop in DocVQA model performance (Table \ref{tab:docvqa_performance}), which is expected due to the increased uncertainty introduced by question rephrasing.

Among the baselines, the $\textsc{SCORE-TA}$ attack proves particularly to be robust, especially against models trained on DocVQA, which show a higher degree of overfitting. In contrast, attacks incorporating loss-based signals introduce additional noise due to uncertainty, leading to a noticeable drop in performance.

Despite the rephrasing, our attacks remain effective, maintaining performance levels comparable to those observed with the original questions, especially against the two PFL models, which demonstrate a lower degree of overfitting.

We also evaluate our proposed attacks against other methods in this setting, focusing on TPR at 1\% and 3\% FPR, with the results summarized in Table \ref{tab:tpr@fpr_rephrased_question_whitebox} and \ref{tab:tpr@fpr_rephrased_question_blackbox}.

\subsection{The resulting Proxy Model}
\label{sec:impact_proxy_model}
The purpose of training the Proxy Model on $D_{\text{query}}$, with labels generated by the black-box model, is to mimic the prediction patterns of the black-box model. The expectation is that the proxy model can capture internal decision-making patterns by following the black-box's prediction strategies. Instead of optimizing for ground-truth labels, we train the proxy to maximize the likelihood of the generated labels. The training process concludes when the proxy achieves near-zero training loss, at which point the final checkpoint is used for the attack.
\begin{figure}[h]
    \centering
    \subfigure[\centering \textbf{Training curve}]{
        \includegraphics[height=2.5cm,width=0.45\textwidth]{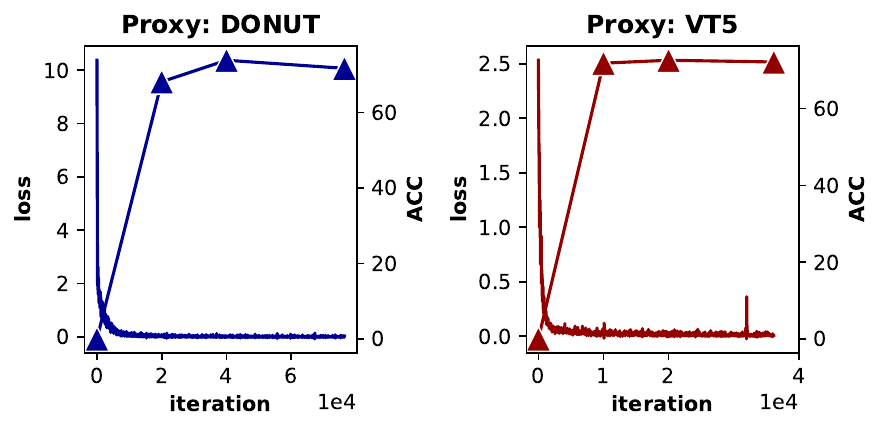}
        \label{fig:proxy_model_train}
    }
    \subfigure[\centering \textbf{Distribution over distance}]{
        \includegraphics[height=2.5cm,width=0.45\textwidth]{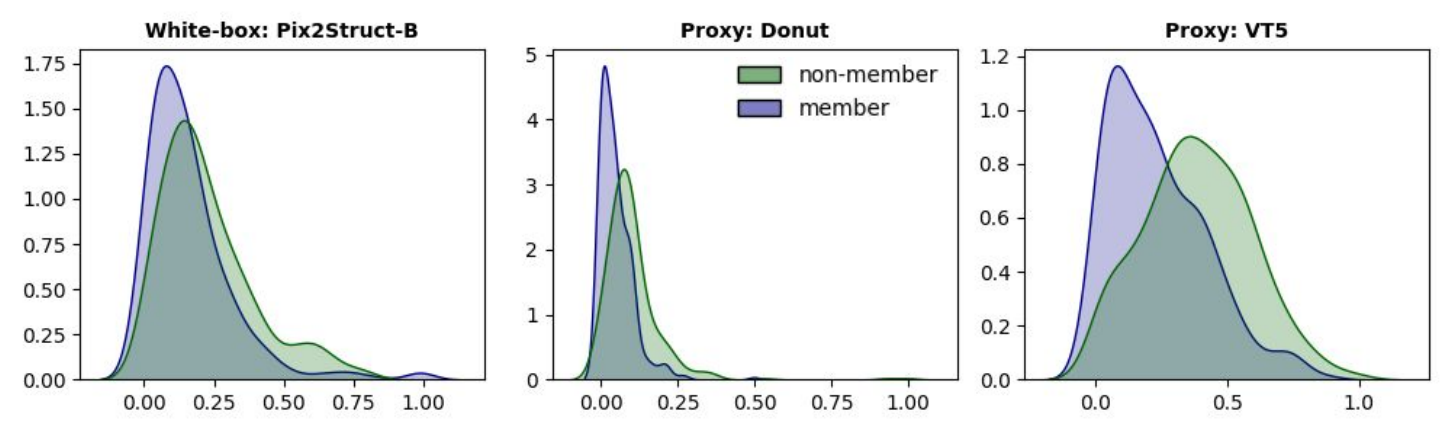}
        \label{fig:proxy_model_distance_distribution}
    }
\caption{\textbf{The resulting Proxy Model} against Pix2Struct-B in the black-box setting. (\textit{a}) The attack accuracy improves quickly once the loss reaches near zero. (\textit{b}) The optimization distance values between member and non-member documents exhibit a separation similar to that seen in the white-box setting.}
\label{fig:proxy_model}
\end{figure}

As illustrated in Figure \ref{fig:proxy_model_train}, the attack performance quickly improves as training progresses. The model overfits quickly, with attack performance reaching its peak early—after just a quarter of the training process—demonstrating the efficiency of our approach. This suggests that \textit{once the proxy model converges, it has effectively captured informative membership signals from the black-box model}, making it ready for the attack. Moreover, we compare the distribution of optimization distances between the proxy model and the same model in the white-box setting, as shown in Figure \ref{fig:proxy_model_distance_distribution}. The results show a similar degree of separation between the two clusters in both cases, indicating the proxy model's effectiveness in approximating the black-box model's behavior to a certain extent.

\subsection{Attack Performance against Minimal-Training Documents}
DocVQA models typically process each question-answer pair independently, resulting in multiple exposures of each document during training. This increases the likelihood of being memorized by the model, making such documents more vulnerable to MIAs. Intuitively, documents associated with fewer training questions should be less exposed and therefore be less vulnerable.

\begin{table}[h]
\begin{center}
\begin{small}
\begin{adjustbox}{width=1\textwidth}
\small
\begin{tabular}{clcccccccc}
\toprule 							
& & \multicolumn{4}{c}{$\textsc{FL}$} & \multicolumn{4}{c}{$\textsc{IG}$}\\
\cmidrule(l){3-6}
\cmidrule(l){7-10}
\multirow{3}{*}{\rotatebox[origin=c]{90}{\textbf{PFL}}} & Model & $m=1$(1) & $m=2$(1) & $m=3$(85) & $\textsc{ALL}$(300) & $m=1$(1) & $m=2$(1) & $m=3$(85) & $\textsc{ALL}$(300)\\
\cmidrule{2-10}
& VT5 & 0 & 0 & 83.53 & 87.67 & 100 & 100 & 85.88 & 86.33\\
& Donut & 100 & 100 & 100 & 97.67 & 100 & 100 & 97.65 & 97\\
\midrule	
\multirow{4}{*}{\rotatebox[origin=c]{90}{\textbf{DVQA}}} & Model & $m=1$(51) & $m=2$(60) & $m=3$(52) & $\textsc{ALL}$(300) & $m=1$(51) & $m=2$(60) & $m=3$(52) & $\textsc{ALL}$(300)\\

\cmidrule{2-10}
& VT5 & 86.27 & 71.67 & 84.62 & 77.00 & 90.2 & 85 & 86.54 & 80.67\\
& Donut & 88.24 & 73.33 & 76.92 & 77.33 & 56.86 & 68.33 & 55.77 & 61.33\\
& Pix2Struct-B & 90.2 & 93.33 & 90.38 & 87 & 88.24 & 88.33 & 76.92 & 73\\		
\bottomrule
\end{tabular}
\end{adjustbox}
\end{small}
\end{center}
\caption{\textbf{Membership Prediction Accuracy on \textit{Member} Documents with minimal repetition.} $m$ denotes the subset of testing documents with $m$ \textit{training} questions, with subset sizes shown in parentheses. Compared to the performance measured on the entire member set (denoted as $\textsc{ALL}$), our attacks are still robust against documents with the low risk of memorization.}
\label{tab:min_repeat_results}
\end{table}

To evaluate this, we measure the accuracy of membership predictions from our attacks on a subset of \textit{member} documents in $D_{\text{test}}$ associated with only a few training questions. These documents represent a minimal memorization risk, posing a more challenging evaluation scenario. Results in Table \ref{tab:min_repeat_results} show that our attacks remain effective on these subsets, achieving high accuracy even for documents $m=1$ training question. This demonstrates the robustness of our attacks under conditions of minimal repetition.

\section{Defenses}

To mitigate the privacy vulnerabilities associated with membership inference attacks in Document Visual Question Answering (DocVQA) systems, we can employ Differential Privacy (DP) techniques~\citep{dwork2014algorithmic}, specifically through the use of differentially private stochastic gradient descent (DP-SGD) introduce by \citet{abadi2016deep}. DP is a robust framework that ensures an individual's data contribution cannot be inferred, even when an adversary has access to the model's outputs. DP-SGD achieves this by adding calibrated noise to the model's gradients during training, thus providing strong theoretical privacy guarantees. However, this approach is not without its drawbacks; the necessity of noise injection can adversely affect the utility of the trained model, leading to reduced performance in answering queries accurately. Alternatively, we can consider ad-hoc solutions such as limiting the number of queries to one question per document in black-box setting, which would inherently reduce the model's usability and flexibility in practical applications. While these measures can enhance privacy, they also necessitate careful consideration of the balance between privacy protection and the functionality of DocVQA systems.

To evaluate the robustness of our proposed membership inference attacks against Differential Privacy (DP), we implemented the well-known DP-SGD algorithm. We considered five privacy budget $\varepsilon \in \{8, 32\}$, with corresponding noise multiplier $\sigma \in \{0.5767822266, 0.3824234009\}$, respectively. 
The composition of the privacy budget over multiple iterations was calculated using Rényi Differential Privacy (RDP). We then converted the RDP guarantees into the standard $(\varepsilon,\delta)$-DP notion following the conversion theorem from \citep{balle2020hypothesis}.

\begin{table}[t]
    \centering
    \begin{minipage}{0.56\textwidth}
        \centering
        \resizebox{\linewidth}{!}{
            \begin{tabular}{lcccccccc}
            \toprule
             & \multicolumn{4}{c}{\textbf{DVQA}} & \multicolumn{4}{c}{\textbf{PFL}}  \\
            \cmidrule(l){2-5}
            \cmidrule(l){6-9}
             & \multicolumn{2}{c}{VT5} & \multicolumn{2}{c}{Donut} & \multicolumn{2}{c}{VT5} & \multicolumn{2}{c}{Donut} \\
            \cmidrule(l){2-3}
            \cmidrule(l){4-5}
            \cmidrule(l){6-7}
            \cmidrule(l){8-9}
             & 1\% & 3\% & 1\% & 3\% & 1\% & 3\% & 1\% & 3\% \\
            \midrule
            Min-K\% & 3.00 & 4.33 & 0.33 & 1.00 & \textbf{6.33} & \textbf{20.33} & 2.00 & 2.33 \\
            Min-K\%++ & 3.00 & 4.67 & 0.00 & 2.67 & 6.33 & 10.00 & 0.00 & 7.00 \\
            \midrule
            FL & 0.67 & 5.00 & \textbf{3.33} & \textbf{8.00} & 3.67 & 17.33 & 3.00 & 4.67 \\
            FLLoRA & \textbf{5.00} & \textbf{9.33} & 0.67 & 3.67 & 5.00 & 9.33 & \textbf{4.33} & \textbf{10.00} \\
            IG & 5.33 & 8.00 & 1.00 & 5.00 & 5.33 & 8.00 & 1.67 & 10.00 \\
            \bottomrule
            \end{tabular}
        }
        \caption{\textbf{White-box Results: TPR at 1\% and 3\% FPR with Rephrased Questions}. Comparison to \textit{white-box} methods: Min-K\% and Min-K\%++ methods, with the best method in \textbf{bold}.}
        \label{tab:tpr@fpr_rephrased_question_whitebox}
    \end{minipage}
    \hfill
    \begin{minipage}{0.43\textwidth}
        \centering
        \resizebox{\linewidth}{!}{
            \begin{tabular}{lcccccc}
             \toprule
             & \multicolumn{2}{c}{\textbf{PFL}} & \multicolumn{4}{c}{\textbf{DVQA}} \\
             \cmidrule(l){2-3}
             \cmidrule(l){4-7}
             & \multicolumn{2}{c}{VT5} & \multicolumn{2}{c}{Donut} & \multicolumn{2}{c}{Pix2Struct-B} \\
             \cmidrule(l){2-3}
             \cmidrule(l){4-5}
             \cmidrule(l){6-7}
             & 1\% & 3\% & 1\% & 3\% & 1\% & 3\% \\
            \midrule
            $\textsc{Score-TA}$& 0.33 & 2.67 & \textbf{3.33} & 9.67 & 3.00 & 8.67 \\
            $\textsc{Score-UA}_{\text{all}}$& 0.33 & 2.67 & 2.33 & 9.33 & \textbf{4.67} & 8.67 \\
            \midrule
            FL & 0.33 & 1.33 & 0.33 & 4.00 & 1.33 & 4.67 \\
            FLLoRA & 1.00 & 5.33 & 1.67 & 5.00 & 2.33 & 6.33 \\
            IG & \textbf{2.67} & \textbf{6.33} & 1.67 & \textbf{11.00} & 3.67 & \textbf{9.33} \\
            \bottomrule
            \end{tabular}
        }
    \caption{\textbf{Black-box Results: TPR at 1\% and 3\% FPR with Rephrased Questions}. Donut is used as The Proxy Model.}
    \label{tab:tpr@fpr_rephrased_question_blackbox}    
    \end{minipage}
    \vskip -0.2in
\end{table}

We trained the Donut model on the DocVQA dataset with DP-SGD to provide theoretical privacy guarantees for individual training documents. Due to resource constraints, we resized document resolution to a smaller size (1280, 960) compared to (2560, 1920) in the public checkpoint  provided by the original authors, which slightly reduced the model's DocVQA performance. For additional details on the effects of document resolution, we refer readers to the original model's paper\citep{Kim22Donut}. The model was trained using the Adam optimizer with a learning rate of $1e-4$, for 10 epochs, and with a batch size of 16. DocVQA performance was evaluated using the Average Normalized Levenshtein Similarity (ANLS) metric.

\begin{table}[h]
\begin{center}
\begin{small}
\begin{adjustbox}{width=1\textwidth}
\small
\begin{tabular}{lcccccccccccc}
\toprule
& \multicolumn{3}{c}{$\varepsilon=8$} & \multicolumn{3}{c}{$\varepsilon=32$} & \multicolumn{3}{c}{$\varepsilon=\infty$} \\
\cmidrule(l){2-4}
\cmidrule(l){5-7}
\cmidrule(l){8-10}
 & ANLS & F1 & TPR@3\%FPR & ANLS & F1 & TPR@3\%FPR & ANLS & F1 & TPR@3\%FPR \\
\midrule
FL & \multirow{3}{*}{19.16} & 55.09 & 2.33 & \multirow{3}{*}{21.81} & 58.84 & 4.33 & \multirow{3}{*}{50.12} & 73.81 & 7.33 \\
FLLoRA &  & 54.94 & 2.00 &  & 58.94 & 3.67 &  & 73.81 & 7.33 \\
IG &  & 56.29 & 1.67 &  & 59.35 & 5.00 &  & 73.52 & 8.67 \\
\bottomrule
\end{tabular}
\end{adjustbox}
\end{small}
\end{center}
\vskip -0.1in
\caption{\textbf{DocMIA Results for Donut trained with DP-SGD on DocVQA dataset}. We report the attack performance of our FL method in terms of F1 score and TPR3\%FPR.}
\label{tab:dp_whitebox}
\end{table}

Table \ref{tab:dp_whitebox} summarizes the results. As expected, introducing DP into model training significantly reduces the attack performance, for example from 73.81\% F1 score with non-DP model to 55.09\% at $\varepsilon = 8$, but this comes at the cost of substantial utility degradation, with the DP model achieving less than half of the performance of the non-DP model, 21.81 of ANLS at $\varepsilon = 8$ compared to 50.12 of ANLS from non-DP checkpoint.
For higher privacy budgets ($\varepsilon = 32$), our attacks demonstrate improved effectiveness, achieving notable gains, +3.75 in F1 and +2 in TPR3\%FPR scores compared to $\varepsilon = 8$, as the model becomes less privacy-constrained.

\end{document}